\documentclass{article}
\usepackage{iclr2027_conference,times}

\usepackage{amsmath,amsfonts,bm}

\def\eqref#1{equation~\ref{#1}}

\def\1{\bm{1}}

\DeclareMathAlphabet{\mathsfit}{\encodingdefault}{\sfdefault}{m}{sl}
\SetMathAlphabet{\mathsfit}{bold}{\encodingdefault}{\sfdefault}{bx}{n}

\newcommand{\softmax}{\mathrm{softmax}}

\newcommand{\KL}{D_{\mathrm{KL}}}

\usepackage{graphicx}
\usepackage{booktabs}
\usepackage{array}
\usepackage{xcolor}
\usepackage{algorithm}
\usepackage{algpseudocode}

\usepackage{float}
\usepackage{placeins}
\usepackage{hyperref}
\hypersetup{pdftitle={DETER: Suppression During Pretraining for Tamper Resistance}, pdfauthor={Maria Ios Glarou, Martin Jaggi}}
\usepackage{url}
\usepackage{xspace}
\usepackage{wrapfig}
\usepackage{enumitem}
\usepackage{caption}
\usepackage{threeparttable}

\newcommand{\deter}{\textsc{Deter}\xspace}
\title{
DETER: Suppression During Pretraining for Tamper Resistance}

\author{%
\makebox[\dimexpr\linewidth-2\tabcolsep\relax][c]{\normalfont
\begin{tabular}{@{}c@{}}
\textbf{Maria Ios Glarou and Martin Jaggi} \\
EPFL \\
\texttt{firstname.lastname@epfl.ch}
\end{tabular}}
}

\iclrfinalcopy
\begin{document}

\maketitle
\lhead{}
\renewcommand{\headrulewidth}{0pt}
\suppressfloats[t]

\begin{abstract}%
Open-weight language models can acquire knowledge that could enable biological, chemical, or cyber misuse. Any attempt to limit such knowledge must remain effective after the model weights are released and can be fine-tuned. Unlearning applied after pretraining can be rapidly undone, whereas filtering the pretraining corpus offers greater resistance. We ask whether actively suppressing knowledge during pretraining can further increase the fine-tuning required to acquire it. Even simple gradient ascent interleaved with standard pretraining on the filtered corpus makes medical knowledge harder to acquire through adversarial fine-tuning than filtering alone or post-hoc unlearning. At 1B parameters, reaching the held-out PubMed loss of an unfiltered model requires over 20$\times$ as many PubMed fine-tuning tokens as after filtering alone. Retained knowledge is evaluated using both held-out data and downstream tasks. Across model sizes, varying suppression strength reveals an increasingly favorable trade-off between fine-tuning difficulty and held-out loss on retained pretraining data. These results point to pretraining as an effective stage for making unwanted capabilities harder to acquire through subsequent fine-tuning. Code is available at \href{https://github.com/MEGlr/DETER}{\texttt{github.com/MEGlr/DETER}}.

\end{abstract}

\section{Introduction}
\label{sec:intro}

Open-weight models pose a particular challenge for machine unlearning: after a model has been modified to suppress knowledge from a target (or \emph{forget}) domain, an adversary with access to the weights can fine-tune it directly on data from that domain~\citep{henderson2023selfdestructing}. \emph{Tamper resistance} measures how well the effect of unlearning persists under such adversarial fine-tuning, or equivalently, how difficult it is to recover the suppressed target-domain capability~\citep{tamirisa2025tamper}. Existing post-training unlearning methods are often brittle under this threat model, with target-domain performance recovering after relatively little additional fine-tuning~\citep{qi2025durability,lee2025distillation,che2025tampering}. Pretraining data filtering offers a substantially more durable alternative~\citep{obrien2026deepignorance}. For knowledge-intensive domains, filtering limits capability acquisition by removing content associated with the target domain before training. The filtering rule must balance reducing target-domain learning against preserving performance on the non-target portion of the training distribution, the \emph{retain distribution}. Finer-grained removal at the token level improves this trade-off~\citep{rathi2026unlearning}, and stronger supervision may come from measuring each document's influence~\citep{yu2024mates} on the target domain relative to the retain distribution. Yet these refinements still estimate which data contribute to target-domain learning, and remain a proxy for what knowledge the final model will ultimately acquire from the retained corpus over the course of pretraining. Target-domain learning can therefore persist even under sophisticated data selection.

\begin{wrapfigure}{r}{0.45\linewidth}
\vspace{-5em}
\centering
\includegraphics[width=\linewidth]{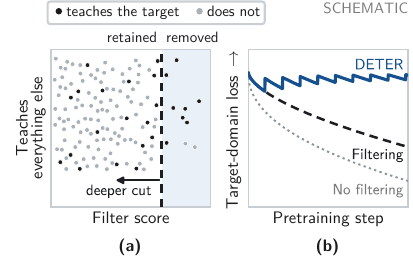}
\caption{\textbf{Conceptual overview.} (a) Filtering decides before pretraining: its selection still contains data that teach the \emph{target domain} along with \emph{retain capabilities}, and a deeper cut removes more of both. (b) Adding periodic suppression provides a second layer of defense against target-domain learning.}
\label{fig:fig1_leak}
\vspace{-2.5em}
\end{wrapfigure}

Filtering provides no explicit signal about how the model should behave on the target domain; its behavior there emerges from training on the retained corpus. The highest-scoring documents removed by filtering can provide an explicit negative training signal, forming a second layer of defense against target-domain learning (Figure~\ref{fig:fig1_leak}). We ask whether actively suppressing target-domain learning during pretraining makes a model more resistant to subsequent target-domain fine-tuning than filtering alone or post-hoc unlearning.

To investigate this question, we pretrain language models of up to 1B parameters from scratch, using medical knowledge as a proxy target domain following \citet{rathi2026unlearning}. Medicine provides a knowledge-intensive domain with measurable capability at these model scales and a closely related retain domain, biology, against which to evaluate the selectivity of suppression. We filter medical documents out of FineWeb-Edu and interleave ordinary pretraining on the retained corpus with suppression updates on a subset of the removed documents. We count how many PubMed fine-tuning tokens it takes to bring each model to the PubMed loss of the unfiltered baseline. At every model size, suppression during pretraining more than doubles this number compared with both filtering alone and Representation Misdirection for Unlearning (\textsc{RMU}; \citealp{li2024wmdp}), a post-hoc unlearning method that steers target-domain representations toward a random vector while preserving representations on retained data (Section~\ref{sec:experiments}).

Crucially, the observed gains are achieved with simple gradient ascent, despite its documented tendency to degrade general capabilities in post-hoc unlearning~\citep{zhang2024npo}. When interleaved with pretraining, this simple objective adds substantial fine-tuning difficulty with limited degradation in retain performance, as measured by corpus and task loss. Just as pretraining lays the foundations for developing useful capabilities through post-training, we argue that it should do the same for making unwanted ones harder to acquire. Building and suppressing capabilities should therefore be studied together as part of base-model design, rather than leaving suppression entirely to interventions after pretraining.

Our primary contributions are:
\begin{itemize}[leftmargin=*,nosep]
    \item We conduct a controlled empirical study of suppressing target-domain knowledge during pretraining on top of document filtering, comparing against filtering alone and post-hoc unlearning across model scales and intervention choices.

    \item We find that suppression during pretraining makes target-domain knowledge substantially harder to acquire through adversarial fine-tuning than filtering alone or post-hoc unlearning, with limited degradation of retained capabilities.

    \item We find favorable scaling behavior in the trade-off between target-domain fine-tuning difficulty and general loss: as model size increases, the suppression frontier shifts toward requiring more fine-tuning at the same general-loss cost.
\end{itemize}

\section{Methodology}
\label{sec:method}

In this section, we describe the setting and how suppression is interleaved with pretraining. Appendix~\ref{app:deter_algorithm} gives the exact training procedure and implementation details.

\paragraph{Setup.} We use \emph{target capability} to refer to a capability we seek to suppress whose acquisition requires exposure to data from a \emph{target domain}, and \emph{retained capabilities} to those we aim to preserve. Let $\mathcal{D}$ denote the pretraining corpus and $\mathcal{D}_F \subseteq \mathcal{D}$ a set of target-domain documents. Let $\mathcal{D}_\rho \subseteq \mathcal{D}$ denote the filtered corpus obtained by removing the documents that a scorer for target-domain content ranks highest, until they account for a fraction $\rho$ of the tokens in $\mathcal{D}$. $\mathcal{D}_F$ is a subset of the removed documents.

Let $\theta_t$ denote the parameters after $t$ pretraining steps, with final parameters $\theta_T$, and let $\bar{\theta}_T$ denote a reference model obtained by standard pretraining on the unfiltered corpus $\mathcal{D}$ for the same number of steps. To measure resistance, an adversary fine-tunes $\theta_T$ on target-domain data. Following \citet{rosati2024immunization}, we measure the fine-tuning compute required to reach the target-domain loss of $\bar{\theta}_T$. We evaluate each model both on resistance and on how well it preserves retained capabilities.

\paragraph{Suppression during pretraining.} Standard pretraining minimizes the next-token prediction loss $\ell_{\mathrm{pre}}$ on $\mathcal{D}_\rho$ with learning rate $\eta_t$. Starting at step $t_{\mathrm{start}}$, we take $S$ suppression steps on $\mathcal{D}_F$ every $P$ pretraining steps (Figure~\ref{fig:suppression_schedule}). For a suppression objective $\mathcal{L}_F$, the suppression steps taken at step $t$ apply
\begin{equation}
\theta^{(s+1)}
=
\operatorname{Update}\!\left(
  \theta^{(s)},
  \nabla_\theta \mathcal{L}_F(\theta^{(s)});
  c\eta_t
\right),
\qquad
\theta^{(0)} = \theta_t,
\quad s=0,\ldots,S-1,
\label{eq:step}
\end{equation}
where $\operatorname{Update}$ denotes one optimizer step, the \emph{suppression strength} $c$ scales the suppression learning rate, and pretraining continues from $\theta^{(S)}$. Suppression is thus specified by the start step $t_{\mathrm{start}}$, the interval $P$, the number of suppression steps $S$, the learning-rate multiplier $c$, and the objective $\mathcal{L}_F$. Unless stated otherwise, $\mathcal{L}_F$ is the negative next-token loss on $\mathcal{D}_F$, that is, gradient ascent \citep{jang2023knowledge}; Section~\ref{sec:discussion} considers other objectives.

\paragraph{Optimizer States.} Pretraining and suppression use the same optimizer types for the same parameter groups and update the same model parameters but maintain separate optimizer states throughout training. For \texttt{AdamW}, pretraining and suppression each maintain their own first- and second-moment estimates. For \texttt{Muon}, they each maintain their own momentum buffer for the attention and MLP weight matrices, together with separate \texttt{AdamW} moment estimates for the parameter groups optimized with \texttt{AdamW}. These quantities are updated only by gradients from the corresponding objective and persist throughout training. Thus, the two objectives interact through the shared model parameters only.

\section{Experiments}
\label{sec:experiments}

To evaluate the effect of suppression during pretraining, we train decoder-only transformers with 113M, 224M, 521M and 1030M parameters from scratch, with medical knowledge as the target domain. The architecture and hyperparameters are detailed in Appendix~\ref{app:models}.

\subsection{Setup}

Following \citet{rathi2026unlearning}, we pretrain on FineWeb-Edu \citep{penedo2024fineweb}. A document classifier ranks its documents by medical content, and we remove the highest-scoring documents up to $\rho = 20\%$ of corpus tokens. $\mathcal{D}_F$ is the set of documents in the top 5\% of corpus tokens. Suppression runs from the start of pretraining ($t_{\mathrm{start}} = 0$): every $P = 50$ pretraining steps, we take a suppression step on $\mathcal{D}_F$ ($S = 1$, $c = 5$). We pretrain and suppress with \texttt{Muon} \citep{jordan2024muon}, using \texttt{AdamW} \citep{loshchilov2019adamw} for the embedding and normalization parameters; $\mu$P transfers one tuned peak learning rate across the four widths \citep{yang2021mup}. We train each model with approximately 20 pretraining tokens per parameter, following the compute-optimal scaling of \citet{hoffmann2022chinchilla}. Experiments were conducted using 1$\times$ H100 80GB or B300 288GB GPU at 113M, 2$\times$ at 224M and 521M, and 5$\times$ A100 80GB at 1030M.

\paragraph{Baselines.} We compare against three baselines: (1) \textsc{Filtering}, pretraining on $\mathcal{D}_\rho$ at $\rho = 20\%$ without suppression; (2) \textsc{RMU} \citep{li2024wmdp}, the post-hoc method most commonly used as a baseline in evaluations of tamper resistance \citep{tamirisa2025tamper, che2025tampering, lee2025distillation, shilov2025sgtm, rathi2026unlearning}, applied to \textsc{Filtering} after pretraining with the same number of target-domain tokens that suppression uses during pretraining; and (3) \textsc{Token filtering} \citep{rathi2026unlearning}, which filters medical content at the token level during pretraining, without suppression. We use its removal variant, which replaces medical tokens with a \texttt{<|hidden|>} token and masks the loss on them. We use a 15\% filtering rate, selected at 113M to match the general loss of suppression and held fixed up to 521M. At 1030M, we use $c = 3$ and a 20\% filtering rate to keep general performance comparable across scales (Table~\ref{tab:resist}). We evaluate token filtering alone as a baseline. Suppression can also be interleaved with token-filtered pretraining, but we leave evaluation of this combination to future work. We report all losses relative to \textsc{No filtering}, i.e., standard pretraining on the full corpus $\mathcal{D}$. Implementation details are provided in Appendix~\ref{app:baselines}.

\subsection{Evaluation}

\begin{figure}
\centering
\includegraphics[width=\linewidth,trim=0 5bp 0 0,clip]{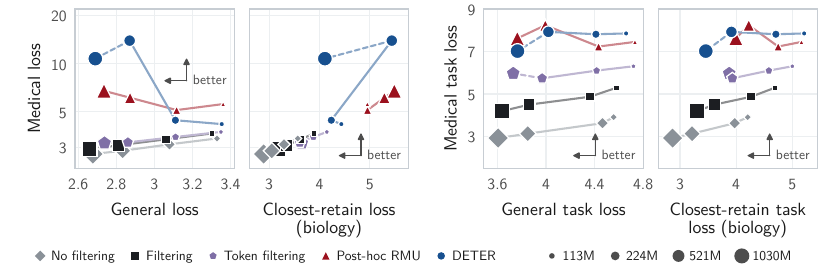}
\caption{
\textbf{At larger scale, suppression separates medicine more strongly from retained capabilities.} Each panel plots medical loss against retained loss before fine-tuning. The first two panels report corpus loss and the last two task loss. For corpus loss, medical, general, and closest-retain (biology) correspond to PubMed, retained FineWeb-Edu, and bioRxiv; for task loss, they correspond to the medical benchmarks, MMLU STEM and non-STEM, and MMLU biology. Larger markers denote larger models. Upward means stronger suppression of medicine; leftward means better preservation of retained capabilities. \textsc{Deter} denotes suppression during pretraining. Dashed segments mark the 1030M settings ($c = 3$; 20\% token filtering).
}
\label{fig:scale_losses}
\vspace{-1.5em}
\end{figure}

\paragraph{Corpus loss.} Following \citet{rathi2026unlearning}, we use held-out cross-entropy loss as a proxy for capability on target and retained domains. We measure medical and general loss on held-out FineWeb-Edu documents labeled by the document classifier: medical (documents scoring at or above the threshold that selects $\mathcal{D}_F$) and general (documents that survive filtering at $\rho = 20\%$, i.e.\ documents in $\mathcal{D}_\rho$). We further report medical loss on PubMed \citep{kandpal2025commonpile}, biology loss on bioRxiv abstracts \citep{muennighoff2023mteb} as the retained domain closest to medicine, and loss on arXiv abstracts \citep{kandpal2025commonpile} and PhilPapers \citep{gao2020pile} to measure broader retained knowledge, following \citet{rathi2026unlearning} (Appendix~\ref{app:corpussets}).

\paragraph{Task Loss.} We complement corpus loss with downstream evaluations chosen to cover the target domain, the closest retained domain, and broader retained knowledge. We evaluate medical knowledge on MedQuAD \citep{benabacha2019medquad}, MedMCQA \citep{pal2022medmcqa}, MedQA-USMLE \citep{jin2021medqa}, and the medicine subsets of MMLU \citep{hendrycks2021mmlu}; biology knowledge on its biology subsets; and general knowledge on its STEM and non-STEM subsets. We report the negative log-likelihood of the correct answer text conditioned on the question (task loss; \citealp{bhagia2025establishing}). This continuous metric provides a more informative signal at small model scales, where discrete accuracy can remain near chance \citep{magnusson2025datadecide}. Appendix~\ref{app:taskloss} details the prompting and loss normalization.

\paragraph{Adversarial fine-tuning.} Following \citet{rathi2026unlearning}, we fine-tune models on the PubMed section of the Common Pile and evaluate on its held-out PubMed corpus loss. We report the number of PubMed fine-tuning tokens required to reach the PubMed corpus loss of \textsc{No filtering}, together with this quantity normalized by the corresponding number required for \textsc{Filtering} at the same model size. The normalized value quantifies the additional fine-tuning required beyond filtering alone. For each model, the adversary uses the learning rate and schedule that reach this threshold with the fewest fine-tuning tokens (Appendix~\ref{app:attack}).

\subsection{Results}

\begin{figure}[H]
\vspace{-1em}
\centering
\includegraphics[width=\linewidth,trim=0 1bp 0 0,clip]{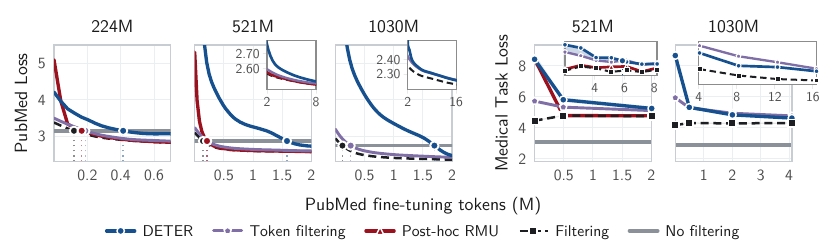}
\caption{
\textbf{Suppression during pretraining slows target-domain learning under fine-tuning on both medical corpus and task loss.} We fine-tune each model on PubMed and report held-out PubMed loss together with medical task loss, averaged over MedQuAD, MedMCQA, MedQA-USMLE, and MMLU medicine. The horizontal line marks the corresponding loss of \textsc{No filtering}; for PubMed, crossing this line defines the fine-tuning budget reported in Table~\ref{tab:resist}. The insets extend the curves to 0.08\% of pretraining. Results are for one fine-tuning seed. \textsc{Deter} denotes suppression during pretraining.
}
\label{fig:pubmed_curves_atkmuon}
\vspace{-1em}
\end{figure}

\paragraph{Suppression degrades medical capability beyond the documents it is applied to.} Suppression is applied only to a subset of the medical documents in the pretraining corpus, yet before any fine-tuning the pretrained model is worse at medicine more broadly: loss increases on held-out PubMed text and on the correct answers of medical benchmarks (Figure~\ref{fig:scale_losses}). The damage is concentrated on medicine and collateral damage is limited. Biology, the closest retained domain, is affected less than medicine, as expected given its overlap with medical text, and general-domain evaluations far less still (Table~\ref{tab:resist}, Appendix Figure~\ref{fig:ga_evals}).

\begin{wrapfigure}{r}{0.35\linewidth}
\vspace{-3.19em}
\centering
\includegraphics[width=\linewidth,trim=0 8.5bp 0 0,clip]{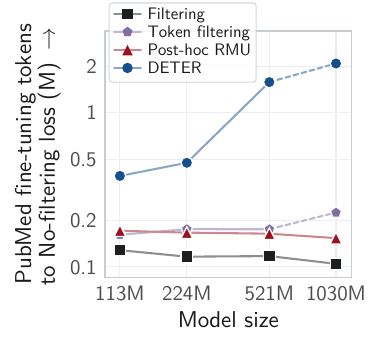}
\caption{
\textbf{Suppression during pretraining requires more target-domain fine-tuning than filtering, token filtering or post-hoc \textsc{RMU}, with the gap growing with model size.} We report the number of PubMed fine-tuning tokens required to reach the PubMed loss of \textsc{No filtering} at each model size. Higher values mean that more fine-tuning is needed to reach the same medical loss. Results are averaged over three fine-tuning seeds. \textsc{Deter} denotes suppression during pretraining. Dashed segments mark the 1030M settings ($c = 3$; 20\% token filtering).}
\label{fig:scale_tokens}
\vspace{-2.29em}
\end{wrapfigure}

\begin{table}[t]
\caption{
\textbf{Suppression during pretraining requires substantially more target-domain fine-tuning than filtering, token filtering or post-hoc \textsc{RMU}, with little change in general loss.} General loss is measured on held-out retained FineWeb-Edu; medical task loss is averaged over MedQuAD, MedMCQA, MedQA-USMLE, and MMLU medicine; general task loss is measured on MMLU STEM and non-STEM; and closest-retain task loss on MMLU biology. All are reported relative to \textsc{No filtering}. The last two columns give the number of PubMed fine-tuning tokens needed to reach the PubMed loss of \textsc{No filtering}, and this number normalized by the corresponding value for \textsc{Filtering} at the same model size. Results are averaged over three fine-tuning seeds. \textsc{Deter} denotes suppression during pretraining. $^\dagger$1030M settings: $c = 3$; 20\% token filtering.}
\small
\setlength{\tabcolsep}{2pt}
\label{tab:resist}
\centering
\setlength{\tabcolsep}{3pt}
\setlength{\belowrulesep}{0.4ex}
\renewcommand{\arraystretch}{0.86}\fontsize{8.5}{9.2}\selectfont
\begin{tabular}{@{}llcccccc@{}}
\toprule
& &
\multicolumn{4}{c}{Before medical fine-tuning on PubMed} &
\multicolumn{2}{c}{\shortstack{After medical fine-tuning}} \\
\cmidrule(lr){3-6}
\cmidrule(l){7-8}
Size & Method
& \shortstack{General\\loss $\downarrow$}
& \shortstack{General\\task loss $\downarrow$}
& \shortstack{Closest-retain\\task loss $\downarrow$}
& \shortstack{Medical\\task loss $\uparrow$}
& \shortstack{Tokens\\(M) $\uparrow$}
& \shortstack{Tokens $\div$\\\textsc{Filtering} $\uparrow$} \\
\midrule
113M
& \textsc{Filtering}       & $-0.03$ & $+0.03$ & $+0.49$ & $+1.39$ & $0.13$ & $1.00\times$ \\
& \textsc{Token filtering} & $+0.02$ & $+0.16$ & $+0.80$ & $+2.39$ & $0.16$ & $1.26\times$ \\
& \textsc{RMU}             & $+0.03$ & $+0.18$ & $+0.96$ & $+3.54$ & $0.17$ & $1.34\times$ \\
& \deter                   & $+0.02$ & $+0.10$ & $+1.01$ & $+3.94$ & $\mathbf{0.39}$ & $\mathbf{3.05\times}$ \\
\midrule
224M
& \textsc{Filtering}       & $-0.02$ & $-0.10$ & $+0.30$ & $+1.24$ & $0.12$ & $1.00\times$ \\
& \textsc{Token filtering} & $+0.03$ & $-0.05$ & $+0.61$ & $+2.47$ & $0.18$ & $1.51\times$ \\
& \textsc{RMU}             & $+0.04$ & $-0.03$ & $+0.78$ & $+3.61$ & $0.17$ & $1.43\times$ \\
& \deter                   & $+0.03$ & $-0.05$ & $+0.73$ & $+4.19$ & $\mathbf{0.47}$ & $\mathbf{4.07\times}$ \\
\midrule
521M
& \textsc{Filtering}       & $-0.02$ & $+0.01$ & $+0.40$ & $+1.37$ & $0.12$ & $1.00\times$ \\
& \textsc{Token filtering} & $+0.03$ & $+0.12$ & $+0.72$ & $+2.61$ & $0.18$ & $1.50\times$ \\
& \textsc{RMU}             & $+0.04$ & $+0.14$ & $+1.01$ & $+5.10$ & $0.16$ & $1.40\times$ \\
& \deter                   & $+0.04$ & $+0.17$ & $+0.70$ & $+4.79$ & $\mathbf{1.58}$ & $\mathbf{13.53\times}$ \\
\midrule
1030M$^\dagger$
& \textsc{Filtering}       & $-0.02$ & $+0.03$ & $+0.44$ & $+1.27$ & $0.10$ & $1.00\times$ \\
& \textsc{Token filtering} & $+0.06$ & $+0.12$ & $+1.01$ & $+3.04$ & $0.23$ & $2.16\times$ \\
& \textsc{RMU}             & $+0.06$ & $+0.16$ & $+1.13$ & $+4.69$ & $0.15$ & $1.47\times$ \\
& \deter                   & $+0.01$ & $+0.16$ & $+0.59$ & $+4.09$ & $\mathbf{2.09}$ & $\mathbf{20.36\times}$ \\
\bottomrule
\end{tabular}
\vspace{-2.4em}
\end{table}
\paragraph{With suppression during pretraining, unwanted knowledge is harder to acquire through adversarial fine-tuning.} Under adversarial fine-tuning on PubMed, suppressed models require substantially more compute than filtering alone or post-hoc \textsc{RMU} to reach the \textsc{No filtering} PubMed loss; \textsc{RMU} uses the same target-token budget and has comparable or higher general loss (Table~\ref{tab:resist}, Appendix~\ref{app:baselines}). Suppression also requires substantially more fine-tuning than token filtering, with the gap widening sharply at 521M (Table~\ref{tab:resist}, Figure~\ref{fig:scale_tokens}). For \textsc{RMU}, medical loss falls steeply early in fine-tuning and approaches the filtering curve, consistent with prior observations for post-hoc unlearning \citep{rathi2026unlearning,shilov2025sgtm} (Figure~\ref{fig:pubmed_curves_atkmuon}). With suppression throughout pretraining, medical loss instead declines more gradually and remains above filtering throughout the attacks we run. For our two largest models, we extend the attack to 0.08\% of pretraining, well beyond the point at which they reach the \textsc{No filtering} loss. On downstream medical tasks, the suppressed model remains above \textsc{Filtering} even at the end of the attack, by a larger margin at 1030M than at 521M, and converges with token filtering by the end of the attack at both sizes (Figure~\ref{fig:pubmed_curves_atkmuon}).

\vspace{-0.4em}
\paragraph{Suppression during pretraining shows scaling behavior.} The fine-tuning compute needed to undo suppression increases more than fivefold from our smallest to our largest model, and the gap relative to token filtering widens with scale (Table~\ref{tab:resist}, Figure~\ref{fig:scale_tokens}). At 1030M, suppression requires $20.36\times$ the fine-tuning of \textsc{Filtering} (2.09M vs.\ 0.10M PubMed tokens). Post-hoc unlearning shows no such trend: \textsc{RMU} stays close to filtering at every size, so at the largest scale it requires nearly fourteen times less fine-tuning than suppression. The added resistance remains concentrated on medicine: at the two largest scales, suppression produces a larger separation between medical and retained loss than the other interventions we test (Figure~\ref{fig:scale_losses}).

\begin{wrapfigure}{l}{0.46\linewidth}
\vspace{-0.5em}
\centering
\includegraphics[width=\linewidth,trim=0 10.5bp 0 0,clip]{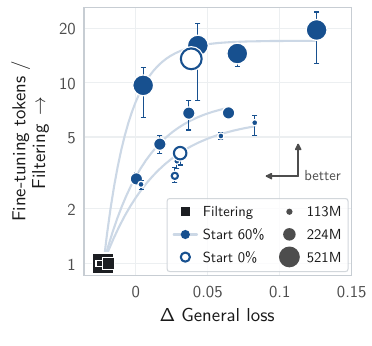}
\caption{
\textbf{At larger scale, suppression requires more target-domain fine-tuning for the same increase in general loss.} General loss is measured on held-out retained FineWeb-Edu, relative to \textsc{No filtering}. Fine-tuning tokens / \textsc{Filtering} is the number of PubMed fine-tuning tokens required to reach the PubMed loss of \textsc{No filtering}, normalized by the corresponding value for \textsc{Filtering} at the same model size. Each curve varies the suppression strength $c$; markers show means over three fine-tuning seeds, and error bars the range. Points toward the upper left give a better trade-off.
}
\label{fig:fig_frontier60}
\vspace{-1em}
\end{wrapfigure}

\paragraph{Starting suppression earlier in pretraining becomes important for selective suppression at scale.} Following \citet{sam2026timing}, we vary the onset of suppression over 0\%, 20\%, and 60\% of pretraining. We keep the number of suppression steps fixed by shortening the interval $P$ for later starts and choosing $c$ to approximately match general loss (Appendix~\ref{app:hparams}). Across start times, later ones yield higher target-domain fine-tuning difficulty (Table~\ref{tab:start_interval}). The shorter interval alone does not account for the difference: using it from the beginning of pretraining does not reproduce the resistance of a late start (Table~\ref{tab:p20_start}). As models scale, however, earlier suppression becomes increasingly favorable for retained capabilities while giving up less of the fine-tuning difficulty achieved by a late start. At 521M, starting from 0\% yields lower general and biology task loss than starting at 60\%, yet retains about 85\% of its normalized fine-tuning difficulty. The same scaling behavior is visible in Figure~\ref{fig:fig_frontier60}: the 0\%-start run lies below the late-start frontier at smaller scales, but approaches it at 521M. This suggests that larger models become amenable to effective suppression earlier in their pretraining trajectory. One possible explanation is that, early in pretraining, the representations supporting medical predictions are not yet sufficiently differentiated from those supporting retained capabilities. Gradient ascent on medical data may therefore disrupt features shared across tasks rather than selectively impair medical capability. Smaller models may require more of their pretraining trajectory to develop representations that permit selective suppression, whereas larger models may reach this regime earlier.

\begin{wraptable}{r}{0.57\linewidth}
\vspace{-2em}
\small
\setlength{\tabcolsep}{3.5pt}
\caption{
\textbf{Starting suppression earlier becomes more effective as model size increases.} We compare suppression onsets while keeping the number of suppression updates fixed and approximately matching general loss. General loss is measured on held-out retained FineWeb-Edu; general task loss on MMLU STEM and non-STEM; and closest-retain task loss on MMLU biology. All are reported relative to \textsc{No filtering} and values near zero suggest better preserved general performance. The final column gives the number of PubMed fine-tuning tokens needed to reach the PubMed loss of \textsc{No filtering}, normalized by the corresponding value for \textsc{Filtering} at the same model size. Results are averaged over three fine-tuning seeds.
}
\label{tab:start_interval}
\centering
\setlength{\tabcolsep}{0.3pt}
\setlength{\aboverulesep}{0.25ex}\setlength{\belowrulesep}{0.4ex}
\begin{tabular}{@{}l@{\hspace{8pt}}lcccc@{}}
\toprule
& &
\multicolumn{3}{c}{\shortstack{Before medical fine-tuning\\on PubMed}} &
\shortstack{After medical\\fine-tuning} \\
\cmidrule(lr){3-5}
\cmidrule(l){6-6}
Size & Start
& \shortstack{General\\loss $\downarrow$}
& \shortstack{General\\task loss $\downarrow$}
& \shortstack{Closest-retain\\task loss $\downarrow$}
& \shortstack{Tokens $\div$\\\textsc{Filtering} $\uparrow$} \\
\midrule
113M & 0\%  & $+0.02$ & $+0.10$ & $+1.01$ & $3.05\times$ \\
     & 20\% & $+0.02$ & $+0.09$ & $+0.90$ & $3.01\times$ \\
     & 60\% & $+0.03$ & $+0.07$ & $+1.18$ & $\mathbf{3.67\times}$ \\
\midrule
224M & 0\%  & $+0.03$ & $-0.05$ & $+0.73$ & $4.07\times$ \\
     & 20\% & $+0.03$ & $-0.08$ & $+0.72$ & $6.11\times$ \\
     & 60\% & $+0.04$ & $+0.03$ & $+0.96$ & $\mathbf{6.78\times}$ \\
\midrule
521M & 0\%  & $+0.04$ & $+0.17$ & $+0.70$ & $13.53\times$ \\
     & 60\% & $+0.04$ & $+0.22$ & $+1.02$ & $\mathbf{16.01\times}$ \\
\bottomrule
\end{tabular}
\vspace{-1em}
\end{wraptable}

\paragraph{Scaling improves the fine-tuning-difficulty--general-loss frontier.} Suppression strength controls the trade-off between resistance to target-domain fine-tuning and retained performance (Figure~\ref{fig:fig_frontier60}). Each curve starts suppression at 60\% of pretraining with $P=20$ and varies only $c$. Along each curve, increasing $c$ initially makes the model require more medical fine-tuning on PubMed to reach the \textsc{No filtering} medical loss, with a relatively small increase in general loss. At larger $c$, this benefit saturates: additional suppression causes substantially more general loss with little further increase in fine-tuning difficulty. As model size increases, this trade-off improves: larger models can require more target-domain fine-tuning for the same increase in general loss. The highest fine-tuning difficulty reached along the curve also increases with model size.

\FloatBarrier
\clearpage
\section{Discussion and limitations}
\label{sec:discussion}

\begin{wrapfigure}{r}{0.4\linewidth}
\vspace{-4em}
\centering
\includegraphics[width=\linewidth,trim=0 12.5bp 0 0,clip]{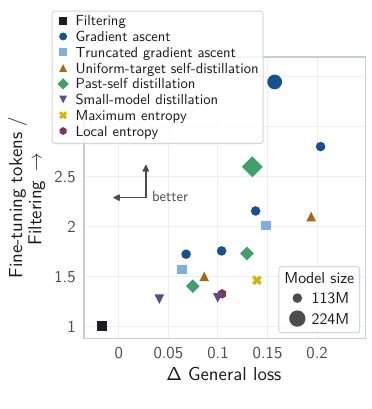}
\caption{
\textbf{Gradient ascent gives the strongest general-loss--fine-tuning difficulty trade-off among the suppression objectives tested.} General loss is measured on held-out retained FineWeb-Edu relative to \textsc{No filtering}. Fine-tuning tokens / \textsc{Filtering} is the number of PubMed fine-tuning tokens required to reach the PubMed loss of \textsc{No filtering}, normalized by the corresponding value for \textsc{Filtering} at the same model size. Points toward the upper left give a better trade-off. Points for the same objective vary the suppression strength $c$. Results are for one fine-tuning seed.
}
\label{fig:fig_pubmed_resist}
\vspace{-1em}
\end{wrapfigure}
\paragraph{Choice of suppression loss.} At 113M with \texttt{AdamW}, we compare gradient ascent with six other choices of $\mathcal{L}_F$. Two carry no target distribution: (1) \emph{truncated gradient ascent} stops the ascent at each token whose probability has fallen below that of a uniform prediction; and (2) \emph{local entropy}, the local loss of PALU \citep{zhai2026palu}, pulls the 5{,}000 largest logits toward the mean logit at the position. Two take as target a distribution in which the observed token, or every token, has the probability of a uniform prediction: (3) \emph{maximum entropy}, which maximises the entropy of the model's next-token distribution; and (4) \emph{uniform-target self-distillation}, the forget term of Unilogit \citep{vasilev2025unilogit}. Two use a weaker predictor of the target domain as the target distribution: (5) \emph{past-self distillation} distils the model's next-token distribution toward that of $\theta_{t_\star}$, an early checkpoint of the same model with $t_\star \le t_{\mathrm{start}}$; and (6) \emph{small-model distillation} distils toward a 61M model pretrained on $\mathcal{D}_\rho$ and held fixed throughout pretraining (see Appendix~\ref{app:losses} for full definitions). Gradient ascent gives the strongest resistance--general-loss trade-off (Figure~\ref{fig:fig_pubmed_resist}). We extend the best-performing objectives to 224M, where the advantage of gradient ascent appears to persist. Gradient ascent lowers the likelihood of observed target-domain tokens without specifying a replacement distribution. Ordinary pretraining, meanwhile, continues to provide a positive learning signal on retained data. One possible explanation for its effectiveness is that the two objectives serve complementary roles: suppression opposes target-domain predictions, while pretraining supports useful predictions elsewhere. More prescriptive objectives add assumptions about what the model should predict on target-domain examples, which appear not to translate into a better trade-off.

\begin{wraptable}{l}{0.57\linewidth}
\centering
\vspace{-1.5em}
\caption{
\textbf{Separate optimizer states add far less closest-retain loss than shared ones, at comparable or greater fine-tuning difficulty.} General loss is measured on held-out retained FineWeb-Edu, medical loss is measured on PubMed, and closest-retain loss on non-medical bioRxiv subjects. All losses are reported relative to \textsc{No filtering}. The final column gives the number of PubMed fine-tuning tokens needed to reach the PubMed loss of \textsc{No filtering}, normalized by the corresponding value for \textsc{Filtering} at the same model size. Results are for one fine-tuning seed.
}
\label{tab:optimizer}
\small
\setlength{\tabcolsep}{3pt}
\begin{tabular}{@{}lcccc@{}}
\toprule
&
\multicolumn{3}{c}{\shortstack{Before medical fine-tuning \\ on PubMed}}
&
\shortstack{After medical\\fine-tuning} \\
\cmidrule(lr){2-4}
\cmidrule(l){5-5}
&
\shortstack{General\\loss $\downarrow$}
&
\shortstack{Closest-retain\\loss $\downarrow$}
&
\shortstack{Medical\\loss $\uparrow$}
&
\shortstack{Fine-tuning\\tokens $\div$\\\textsc{Filtering} $\uparrow$}\\
\midrule

\multicolumn{5}{@{}l}{\texttt{Muon}, 113M} \\
Separate & $-0.02$ & $+0.57$  & $+0.47$  & \textbf{1.66$\times$} \\
Shared   & $-0.01$ & $+25.60$ & $+10.99$ & 1.62$\times$ \\

\midrule
\multicolumn{5}{@{}l}{\texttt{AdamW}, 113M} \\
Separate & $+0.07$ & $+0.75$  & $+0.75$  & \textbf{1.93$\times$} \\
Shared   & $+0.05$ & $+14.13$ & $+13.83$ & 1.79$\times$ \\

\midrule
\multicolumn{5}{@{}l}{\texttt{AdamW}, 224M} \\
Separate & $+0.09$ & $+2.02$  & $+8.11$  & \textbf{3.35$\times$} \\
Shared   & $+0.09$ & $+54.10$ & $+25.23$ & 2.47$\times$ \\
\bottomrule
\end{tabular}
\vspace{-1em}
\end{wraptable}

\paragraph{How should suppression interact with pretraining?} All models in our main experiments maintain separate optimizer statistics for pretraining and suppression (Section~\ref{sec:method}). Sharing statistics makes suppression appear stronger at the end of pretraining, but nearly all of the additional medical loss disappears within the first few adversarial fine-tuning steps. It also causes much greater spillover into biology despite comparable general loss. Separate statistics instead largely avoid this cost and, in the larger model, also yield greater resistance (Table~\ref{tab:optimizer}). Sharing statistics couples the objectives beyond their effects on model parameters: pretraining gradients influence the optimizer history used for suppression, and suppression gradients influence the history used for pretraining. Separate statistics keep those histories distinct while allowing both objectives to shape the same model. For \texttt{AdamW}, the distinction extends beyond momentum: at 113M keeping separate first moments while sharing second moments severely degrades general performance. Independent momentum histories alone therefore do not reproduce the benefits of fully separate optimizer states.

\begin{wrapfigure}{l}{0.45\linewidth}
\centering
\vspace{-1.5em}
\includegraphics[width=\linewidth]{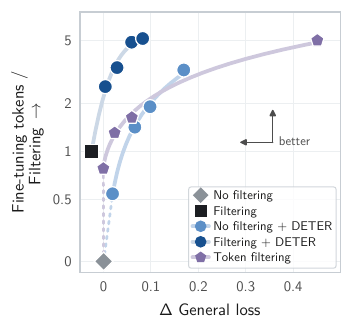}
\caption{
\textbf{Suppression increases the fine-tuning required even without filtering, and is strongest when combined with document filtering.} We plot general loss relative to \textsc{No filtering} against the PubMed fine-tuning tokens required to reach the loss of \textsc{No filtering}, normalized by \textsc{Filtering}. For suppression with and without document filtering, connected points vary the suppression strength $c$, with suppression starting at 60\% of pretraining and $P=20$; for \textsc{Token filtering}, they vary the filtering rate. \textsc{No filtering} and \textsc{Filtering} are shown as reference points. Points toward the upper left are better. Results are at 113M, for one fine-tuning seed. \textsc{Deter} denotes suppression during pretraining.
}
\label{fig:fig_113_frontier}
\vspace{-2em}
\end{wrapfigure}

\paragraph{Does suppression require filtering?}
At 113M, suppression increases fine-tuning difficulty even without filtering, but combining it with document filtering yields a substantially better trade-off with general loss (Figure~\ref{fig:fig_113_frontier}). Filtering may become more important at scale. Without filtering, standard pretraining encourages learning medical knowledge, whereas suppression applies an opposing gradient signal. A larger model might have sufficient capacity to accommodate both objectives, retaining medical knowledge despite suppression. Filtering reduces this conflict by removing medical content from standard pretraining. Token filtering goes further by removing medical spans from documents retained by document filtering, potentially becoming more important as models scale. Consistent with this possibility, token filtering catches up with suppression earlier on medical tasks at 1030M than at 521M (Figure~\ref{fig:pubmed_curves_atkmuon}). This motivates applying suppression on top of token filtering.

\paragraph{At what granularity should suppression act?} At matched general loss, restricting suppression to medical tokens selected by a token-level mask (Appendix~\ref{app:mask}) reduces spillover to biology, but lowers resistance relative to suppressing the full target-domain document (Table~\ref{tab:positions}). The comparison is made at a scale where the suppression steps can draw entirely from the most medical portion of the filtered-out data, where target-domain content is already highly concentrated. The weaker resistance of token-level suppression may partly reflect noise in the token-level mask, rather than an inherent disadvantage of operating at finer granularity. At larger pretraining scales, the additional suppression updates could instead be used to broaden coverage of the target domain by drawing on more of the filtered-out corpus. Because these additional documents contain a greater mixture of medical and retain-domain content, token-level selection may prove useful in that regime.

\paragraph{Which data should suppression use?} The data most useful for suppression need not be identical to the data that should be removed from pretraining. We construct $\mathcal{D}_F$ from the highest-scoring FineWeb-Edu documents accounting for 5\% of corpus tokens, within the broader 20\% removed by filtering. This uses different selection thresholds for excluding potential sources of target-domain learning and supplying negative training examples. A high classifier score, however, does not establish that a document provides the most useful suppression signal or that the selected set covers the full target domain. A narrow selection may leave some subdomains underrepresented, whereas a broader selection may introduce more retain-domain content. The observed transfer to PubMed and medical benchmarks demonstrates effects beyond the suppression documents, but does not establish that the current selection is optimal. Comparing selection thresholds, subdomain coverage, and alternative sources of suppression data would clarify their effects on resistance and collateral damage.

\paragraph{How do classifier errors affect suppression?} Classifier errors affect both the pretraining stream and the suppression signal. False negatives leave target-domain documents available for ordinary pretraining. Because these documents are absent from the suppression set, suppression can counteract learning from them only insofar as its effects generalize beyond that set. False positives in filtering remove potentially useful retain-domain content; those also selected for suppression additionally train the model against that content, creating a further route to collateral damage. Classifier quality therefore determines not only what the model learns from, but also what it is trained against.

\begin{wraptable}{r}{0.48\linewidth}
\centering
\vspace{-1.5em}
\caption{
\textbf{Token-level suppression makes capability unlearning more concentrated, but document-level suppression makes target capability harder to acquire with fine-tuning.} All losses are relative to \textsc{No filtering}; retain/target is the increase on bioRxiv divided by that on PubMed. The final column gives the number of PubMed fine-tuning tokens needed to reach the PubMed loss of \textsc{No filtering}, normalized by the corresponding value for \textsc{Filtering} at the same model size. Results are averaged over three fine-tuning seeds. Token-level suppression applies the loss only to the tokens that a token-level probe labels medical; the confident variant applies it only to the tokens the probe scores highest.
}
\label{tab:positions}
\footnotesize
\setlength{\tabcolsep}{2.5pt}
\begin{tabular}{@{}lccc@{}}
\toprule
& \shortstack{General\\loss $\downarrow$}
& \shortstack{Retain /\\target $\downarrow$}
& \shortstack{Tokens $\div$\\\textsc{Filtering} $\uparrow$} \\
\midrule
\multicolumn{4}{@{}l}{\texttt{AdamW}, 113M} \\
\textsc{Filtering}   & $-0.02$ & 1.41 & $1.00\times$ \\
Token, confident     & $+0.07$ & \textbf{1.01} & $1.81\times$ \\
Token                & $+0.07$ & 1.03 & $1.96\times$ \\
Document             & $+0.07$ & 1.08 & $\mathbf{2.00\times}$ \\
\midrule
\multicolumn{4}{@{}l}{\texttt{Muon}, 113M} \\
\textsc{Filtering}   & $-0.03$ & 1.29 & $1.00\times$ \\
Token, confident     & $+0.02$ & \textbf{1.04} & $2.43\times$ \\
Token                & $+0.03$ & 1.12 & $2.94\times$ \\
Document             & $+0.02$ & 1.11 & $\mathbf{3.05\times}$ \\
\bottomrule
\end{tabular}
\vspace{-1em}
\end{wraptable}

\paragraph{Limitations.} Our experiments use medical knowledge as a proxy target domain in models of up to 1030M parameters, and therefore do not establish that the observed scaling trends extend to genuinely hazardous capabilities or to other domains and bigger model scales. We do not systematically vary classifier quality or suppression-set construction, leaving sensitivity to false positives, missed subdomains, and the source and diversity of suppression data unresolved. The same classifier also defines the FineWeb-Edu evaluation splits, which may share blind spots with training-data selection; the external corpora and benchmarks provide complementary but not exhaustive coverage. Furthermore, recovery is measured through held-out PubMed loss under fine-tuning attacks on models pretrained with a single seed; we do not exhaustively evaluate adversarial modifications. Every method reaches the \textsc{No filtering} baseline's PubMed loss within the evaluated attack budget. In absolute terms, these budgets are small: at the largest scale, filtering reaches the \textsc{No filtering} loss after 0.10M PubMed tokens and suppression after 2.09M, or 0.010\% of its pretraining tokens. Our findings therefore establish that more target-domain fine-tuning is required under the attacks and the model sizes we test.

\paragraph{Future work.} Several directions would extend the present study. Suppression could also be interleaved with token-filtered pretraining. Broader comparisons could evaluate \textsc{SGTM}~\citep{shilov2025sgtm} and \textsc{TAR}~\citep{tamirisa2025tamper}, trace filtering and suppression trade-offs more extensively, account for the additional suppression compute when comparing retained utility, and test whether using different optimizers for pretraining and suppression increases the amount of target-domain fine-tuning required while preserving retained performance, as explored for separate objectives in post-hoc unlearning~\citep{zhong2025dualoptim}. Extending the experiments to larger models and additional target domains would test how broadly the observed scaling behavior holds and enable more controlled study of the factors underlying it. Finally, representation-level analyses could investigate why suppression during pretraining makes target-domain capability slower to acquire through fine-tuning, including whether interventions differ in residual target knowledge or in the organization of target and retained representations.

\section{Conclusion}
\label{sec:conclusion}
The effectiveness of simple gradient ascent motivates studying the fundamentals of suppression within pretraining. At larger scale, earlier suppression improves the balance between resistance and general loss, highlighting the importance of how suppression interacts with ordinary learning. Target-domain learning through fine-tuning remains possible; the goal is to raise its cost while preserving useful capabilities. Establishing how to achieve that balance across scales and domains should be part of understanding pretraining itself, rather than left entirely to the study of post-hoc unlearning.

\FloatBarrier
\clearpage

\subsection*{AI use statement}
We used AI tools to assist with code implementation and to provide feedback on research methodology and experiments. We additionally used these tools for manuscript editing and literature search.

\subsection*{Reproducibility statement}
Section~\ref{sec:method} describes the suppression procedure. Appendix~\ref{app:implementation} provides model architectures, optimization details, and the full training procedure; Appendix~\ref{app:hparams} reports the suppression hyperparameters used in each experiment; Appendix~\ref{app:data} describes data construction and target-domain selection; and Appendix~\ref{app:eval} gives baseline implementations and evaluation details, including the adversarial fine-tuning procedure and hyperparameter search. Code is available at \url{https://github.com/MEGlr/DETER}.

\bibliography{iclr2027_conference}
\bibliographystyle{iclr2027_conference}

\clearpage

\appendix

\section{Related work}
\label{app:related}
\suppressfloats[t]

\paragraph{Preference-informed pretraining.} \citet{korbak2023pretraining} incorporate human preferences into pretraining to steer model generations away from undesirable behavior, finding conditional training particularly effective. Their results show that safety-relevant signals can be more effective when introduced during pretraining than through subsequent fine-tuning. We study a different objective: suppressing the acquisition of domain knowledge and increasing the fine-tuning required to acquire it later.

\paragraph{Pretraining data filtering.} Filtering suppresses capabilities by removing their training signal altogether. \citet{obrien2026deepignorance} show that filtering biothreat-related data can remain effective under extensive adversarial fine-tuning, and \citet{rathi2026unlearning} find that token-level filtering improves the retain--forget trade-off, strengthens with scale, and remains effective under label noise given sufficient pretraining compute. We use filtering as our primary baseline and ask whether the excluded target-domain data can instead be used to make the resulting model harder to fine-tune on that domain.

\paragraph{Intervention timing.} The effect of a pretraining intervention can depend on when it begins. For behavior-shaping safety interventions, \citet{sam2026timing} find that earlier introduction generally improves robustness to subsequent fine-tuning; for capability filtering, \citet{rathi2026unlearning} find that delaying loss masking weakens suppression. We therefore treat suppression start time as an experimental variable.

\paragraph{Localizing capabilities during training.} Gradient Routing and Selective GradienT Masking (\textsc{SGTM}) control where target-domain information is learned by restricting its gradients to designated parameters that can later be removed \citep{cloud2024gradientrouting,shilov2025sgtm}. \textsc{SGTM} is robust to adversarial fine-tuning and can outperform filtering when target-domain labels are imperfect. Our approach uses the shared model parameters throughout, testing whether periodic suppression updates can make target knowledge harder to acquire without reserving or removing parameters.

\paragraph{Post-hoc unlearning.} Post-hoc unlearning modifies a model after pretraining using objectives on its outputs \citep{jang2023knowledge,zhang2024npo,fan2025simnpo,zhai2026palu,vasilev2025unilogit} or representations \citep{li2024wmdp}. Such interventions can be rapidly reversed by fine-tuning \citep{lee2025distillation}. Optimizer choice and state can also affect post-hoc unlearning: \citet{zhong2025dualoptim} use a separate adaptive optimizer for forgetting and maintain separate optimizer states for forget and retain updates. In our setting, pretraining and suppression use the same optimizer types for the same parameter groups, and we study whether their optimizer states should be shared. We compare against post-hoc \textsc{RMU} to test whether applying suppression during the pretraining trajectory changes the subsequent cost of acquiring target-domain knowledge.

\paragraph{Tamper resistance.} Tamper-resistant training explicitly optimizes models against simulated fine-tuning adversaries \citep{henderson2023selfdestructing,tamirisa2025tamper}. Its evaluation depends critically on attack strength: \citet{qi2025durability} show that stronger fine-tuning configurations can substantially reduce previously reported durability. Following \citet{rathi2026unlearning}, we therefore tune the adversarial fine-tuning learning rate and measure the target-domain fine-tuning required to reach the corresponding \textsc{No filtering} model.

\section{Implementation details}
\label{app:implementation}

\subsection{Models and pretraining}
\label{app:models}

We follow the architecture and scale ladder of \citet{rathi2026unlearning}: decoder-only transformers with rotary position embeddings, squared-ReLU activations, pre-RMSNorm, a width-to-depth ratio of 64, and $\mu$P transfer from a base width of 512. All models use a context length of 2,048 tokens and an effective batch size of 327,680 tokens. Following \citet{rathi2026unlearning}, we use a peak learning rate of $3{\times}10^{-3}$ at the base width for the weight matrices inside the transformer blocks and transfer it across scales with $\mu$P. The learning-rate schedule uses linear warmup over the first 10\% of training followed by cosine decay to 10\% of the peak rate. Each model is trained Chinchilla-optimally, on 20 tokens per parameter \citep{hoffmann2022chinchilla}.

We pretrain models under two optimizer configurations, which we refer to by their main optimizer. Under \texttt{Muon}, we update the weight matrices inside the transformer blocks with \texttt{Muon} and the embedding and normalization parameters with \texttt{AdamW}. We rescale each orthogonalized update to match the root-mean-square of an \texttt{AdamW} update \citep{liu2025moonlight}, and orthogonalize the query, key, and value projections separately. Under \texttt{AdamW}, we update all parameters with \texttt{AdamW}.

The unfiltered baseline, filtered baseline, and suppression-pretrained models use identical architecture and optimization settings and differ only in the suppression configuration described in Section~\ref{sec:method}. Table~\ref{tab:arch} reports the architecture at each scale.

\begin{table}[H]
\caption{Architecture for each model scale.}
\label{tab:arch}
\begin{center}
\begin{tabular}{@{}lrrr@{}}
\toprule
Model & Layers & Heads & Embed dim \\
\midrule
61M  & 7  & 8  & 448 \\
113M & 10 & 10 & 640 \\
224M & 14 & 14 & 896 \\
521M & 20 & 10 & 1,280 \\
1030M & 26 & 16 & 1,664 \\
\bottomrule
\end{tabular}
\end{center}
\end{table}

\subsection{Suppression steps}
\label{app:deter_algorithm}
Algorithm~\ref{alg:deter} gives the training procedure. Pretraining and suppression steps apply the same optimizer to the same parameters $\theta$, but each keeps its own gradient statistics. $s$ and $s_F$ hold what the optimizer accumulates from past gradients: for momentum and matrix-based methods (\texttt{SGD} with momentum, \texttt{Muon}), one momentum buffer per parameter; for coordinate-wise adaptive methods (\texttt{Adam}, \texttt{AdamW}), the first- and second-moment estimates. $s$ is updated only by gradients of $\ell_{\mathrm{pre}}$, and $s_F$ only by gradients of $\mathcal{L}_F$ as in Eq.~(\ref{eq:step}) (Figure~\ref{fig:separate_statistics}).

\begin{figure}[H]
\centering
\includegraphics[width=\linewidth]{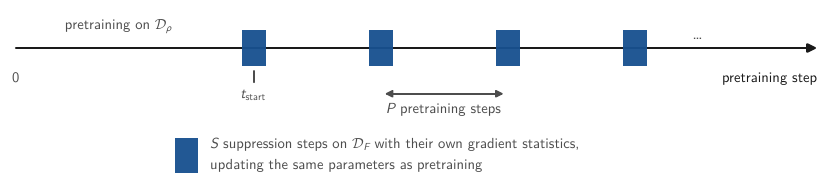}
\caption{Schedule of suppression steps during pretraining.}
\label{fig:suppression_schedule}
\end{figure}

\begin{algorithm}[htp]
\caption{Suppression during pretraining.}
\label{alg:deter}
\begin{algorithmic}[1]
\Require $\mathcal{D}_\rho$: filtered corpus; $\mathcal{D}_F \subseteq \mathcal{D}$: target-domain documents
\Require $\ell_{\mathrm{pre}}$: next-token loss; $\mathcal{L}_F$: suppression objective
\Require $T$: pretraining steps; $t_{\mathrm{start}}$: first suppression step; $P$: interval; $S$: suppression steps
\Require $c$: suppression learning-rate multiplier
\Require $\mathrm{Opt}$: optimizer, with learning-rate schedule $\eta_t$ and weight decay $\lambda$
\Require $s$, $s_F$: gradient statistics of pretraining and of suppression
\Require $\theta_0$: initial parameters
\State $\theta \leftarrow \theta_0$, \quad $s, s_F \leftarrow 0$
\For{$t = 0$ \textbf{to} $T-1$}
  \If{$t \ge t_{\mathrm{start}}$ \textbf{and} $t \bmod P = 0$} \hfill\textcolor{gray}{\footnotesize \# \textit{Suppression step}}
    \For{$j = 1$ \textbf{to} $S$}
      \State Sample a batch $B_F \sim \mathcal{D}_F$
      \State Compute the suppression loss $\mathcal{L} \leftarrow \mathcal{L}_F(\theta; B_F)$
      \State Update parameters $(\theta, s_F) \leftarrow \mathrm{Opt}\!\left(\theta,\, \nabla_\theta \mathcal{L},\, s_F;\, c\,\eta_t,\, \lambda\right)$
    \EndFor
  \EndIf
  \State Sample a batch $B \sim \mathcal{D}_\rho$ \hfill\textcolor{gray}{\footnotesize \# \textit{Pretraining step}}
  \State Update parameters $(\theta, s) \leftarrow \mathrm{Opt}\!\left(\theta,\, \nabla_\theta \ell_{\mathrm{pre}}(\theta; B),\, s;\, \eta_t,\, \lambda\right)$
\EndFor
\State \textbf{Return} parameters $\theta_T$
\end{algorithmic}
\end{algorithm}

\paragraph{Efficiency analysis.} When using the same batch size, a suppression update requires approximately the same compute as a pretraining update: one forward and one backward pass. During the interval in which suppression is active, the additional compute is therefore approximately $S/P$ of pretraining compute. Losses requiring an additional forward pass through a reference model of the same size increase this estimate to $4S/(3P)$. For $S=1$ and $P=50$, these correspond to approximately $2\%$ and $2.7\%$, respectively. A delayed suppression start reduces the overhead averaged over the full pretraining. The primary memory overhead stems from the second set of moment estimates that suppression maintains. If model parameters require $M$ storage, \texttt{AdamW} stores approximately $4M$ (parameters, gradients, and 2 EMA states). The suppression steps store additionally $2M$ for the two EMA states, leading to $\approx 50\%$ memory overhead. With \texttt{Muon}, the weight matrices inside the transformer blocks carry a single momentum buffer rather than two moment estimates, halving their contribution to this overhead, which tends toward $\approx 33\%$ as transformer parameters dominate at larger scales. Losses requiring a reference model need additional memory for the reference model's parameters and gradients.

\begin{figure}[t]
\begin{center}
\includegraphics[width=\linewidth]{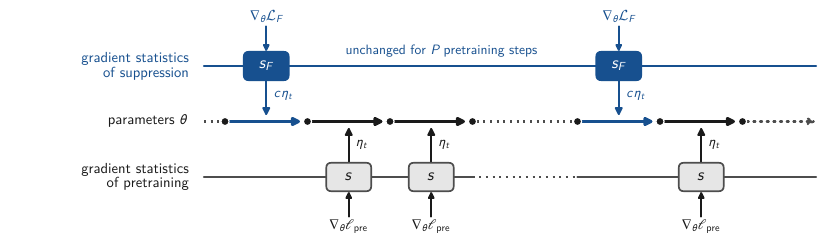}
\end{center}
\caption{Optimization states of pretrainer (learner) and suppressor (unlearner) throughout pretraining with interleaved suppression.}
\label{fig:separate_statistics}
\end{figure}

\subsection{Suppression losses}
\label{app:losses}

We specify the per-token suppression loss $\ell$ in Eq.~(\ref{eq:step}) for each loss considered in Section~\ref{sec:discussion}. All losses below are minimized in the suppression update; in particular, minimizing $\log p_\theta(x_i\mid x_{<i})$ corresponds to gradient ascent on the next-token cross-entropy. Let $z_\theta(x_{<i})\in\mathbb{R}^{|V|}$ denote the logits at position $i$, $p_\theta(\cdot\mid x_{<i})$ their softmax, and $x_i$ the observed token. Teacher and target distributions are treated as fixed during each suppression update.

\begin{itemize}[leftmargin=*]
\item \textbf{Gradient ascent.} Following \citet{jang2023knowledge},

$$ \ell(\theta;x,i) = \log p_\theta(x_i\mid x_{<i}). $$

\item \textbf{Truncated gradient ascent.} We stop suppressing a token once its probability falls below that of a uniform prediction:

$$ \ell(\theta;x,i) = \max\!\left\{ \log p_\theta(x_i\mid x_{<i}),-\log |V| \right\}. $$

\item \textbf{Local entropy.} Following the local-entropy objective of PALU \citep{zhai2026palu}, we flatten the $K$ largest logits toward the mean logit at each selected position:

$$ \ell(\theta;x,i) = \frac{1}{K} \sum_{v\in V_{\mathrm{top}}} \left(z_{\theta,v}(x_{<i})-\bar z\right)^2, \qquad \bar z = \frac{1}{|V|} \sum_{v\in V}z_{\theta,v}(x_{<i}). $$

We use $K=5{,}000$ and, unlike PALU, recompute $V_{\mathrm{top}}$ from the current model at each suppression step. The loss is evaluated at every position selected by $m$.

\item \textbf{Maximum entropy.} We minimize the divergence from a uniform next-token distribution:

$$ \ell(\theta;x,i) = \log |V|-H\!\left(p_\theta(\cdot\mid x_{<i})\right) = \KL\!\left( p_\theta(\cdot\mid x_{<i}) \,\big\|\, \mathrm{Unif}(V) \right). $$

\item \textbf{Uniform-target self-distillation.} We use the forget objective of Unilogit \citep{vasilev2025unilogit}. At each suppression step, we construct a detached target from the current logits, changing only the logit of $x_i$ so that its target probability equals $1/|V|$:

$$ \ell(\theta;x,i) = \KL\!\left( p_\theta(\cdot\mid x_{<i}) \,\big\|\, \softmax(\tilde z) \right), $$

where

$$ \tilde z_v =
\begin{cases}
\displaystyle \log \frac{\sum_{u\neq x_i}e^{z_{\theta,u}(x_{<i})}}
{|V|-1},
& v=x_i,\\[6pt]
z_{\theta,v}(x_{<i}),
&v\neq x_i.
\end{cases}
$$

The target $\tilde z$ is detached before computing the KL divergence.

\item \textbf{Past-self distillation.} We distil toward a frozen checkpoint $\theta_{t_\star}$ of the same model, with $t_\star\leq t_{\mathrm{start}}$:

$$ \ell(\theta;x,i) = \KL\!\left( p_\theta(\cdot\mid x_{<i}) \,\big\|\, p_{\theta_{t_\star}}(\cdot\mid x_{<i}) \right). $$ We use the checkpoint after 5\% of pretraining as the teacher. Early in pretraining, language models predict mainly from low-order statistics such as token and $n$-gram frequencies \citep{belrose2024statistics, chang2024learningcurves}. Distilling toward such a teacher replaces target-domain predictions with generic ones that pretraining itself produces, rather than with the uninformative predictions of a teacher at initialization. At matched general loss, the 5\% teacher gives higher resistance (Table~\ref{tab:past_self_checkpoint}). We do not tune the checkpoint further. We also evaluate a \emph{rolling teacher}, updated recursively: after each suppression update, the resulting model replaces the teacher. Unlike a fixed early checkpoint, a rolling teacher tracks the general capabilities of the model throughout pretraining, while carrying forward the effect of earlier suppression updates. It is, however, not guaranteed to remain a weak predictor of the target domain: target-domain knowledge that survives a suppression update propagates to all subsequent teachers, a risk we expect to increase under weaker filtering. Empirically, the rolling teacher performs comparably to the frozen 5\% checkpoint (Table~\ref{tab:past_self_checkpoint}).

\item \textbf{Small-model distillation.} We distil toward a frozen 61M model with the same vocabulary:

$$ \ell(\theta;x,i) = \KL\!\left( p_\theta(\cdot\mid x_{<i}) \,\big\|\, p_{\mathrm{small}}(\cdot\mid x_{<i}) \right). $$

We consider three teachers: a model pretrained on $\mathcal{D}_\rho$ with $\rho=20\%$; the same model trained with gradient-ascent suppression; and a model pretrained on $\mathcal{D}$ and subsequently unlearned with RMU \citep{li2024wmdp}.

\end{itemize}

\begin{table}[H]
\caption{Past-self distillation at 113M with \texttt{AdamW}, with the teacher at initialization (0\%), after 5\% of pretraining, and with a rolling teacher. General loss is measured on held-out retained FineWeb-Edu, relative to \textsc{Filtering}. The final column gives the number of PubMed fine-tuning tokens needed to reach the PubMed loss of \textsc{No filtering}, normalized by the corresponding value for \textsc{Filtering}. Results are for one fine-tuning seed.}
\label{tab:past_self_checkpoint}
\begin{center}
\begin{tabular}{@{}l c c@{}}
\toprule
Teacher & General loss $\downarrow$ & Tokens $\div$ \textsc{Filtering} $\uparrow$ \\
\midrule
0\% & $+0.08$ & $1.04\times$ \\
5\% & $+0.09$ & $1.42\times$ \\
Rolling & $+0.10$ & $1.45\times$ \\
\bottomrule
\end{tabular}
\end{center}
\end{table}

\paragraph{Retain loss.} Unlearning methods typically add a retain loss, which keeps the model close to a frozen copy of itself on general data \citep{li2024wmdp, vasilev2025unilogit, zhai2026palu}. In our setting, the retain loss is a KL divergence on general documents to the predictions of $\theta_t$, the model just before the suppression update; when $S \geq 2$, it adds a forward pass of $\theta_t$ and a forward and backward pass at every suppression step after the first. Since we take one or very few suppression steps, and the pretraining steps between them protect general capabilities, we drop the retain loss to avoid its cost and let $c$ and $P$ control suppression (Table~\ref{tab:retain}).

\begin{table}[H]
\caption{Past-self distillation at 113M with \texttt{AdamW}, with and without the retain loss. General loss is measured on held-out retained FineWeb-Edu, relative to \textsc{Filtering}. Tokens $\div$ \textsc{Filtering} is the number of PubMed fine-tuning tokens needed to reach the PubMed loss of \textsc{No filtering}, normalized by the corresponding value for \textsc{Filtering}. Compute is the cost of one suppression in pretraining steps. Results are for one fine-tuning seed.}
\label{tab:retain}
\begin{center}
\begin{tabular}{@{}l c c c c@{}}
\toprule
Retain loss & $S$ & General loss $\downarrow$ & Tokens $\div$ \textsc{Filtering} $\uparrow$ & Compute $\downarrow$ \\
\midrule
No  & 1 & $+0.09$ & $1.42\times$ & 1 \\
Yes & 2 & $+0.09$ & $1.43\times$ & $3\tfrac{1}{3}$ \\
No  & 2 & $+0.13$ & $1.51\times$ & 2 \\
\bottomrule
\end{tabular}
\end{center}
\end{table}

\begin{figure}[H]
\begin{center}
\includegraphics[width=0.7\linewidth]{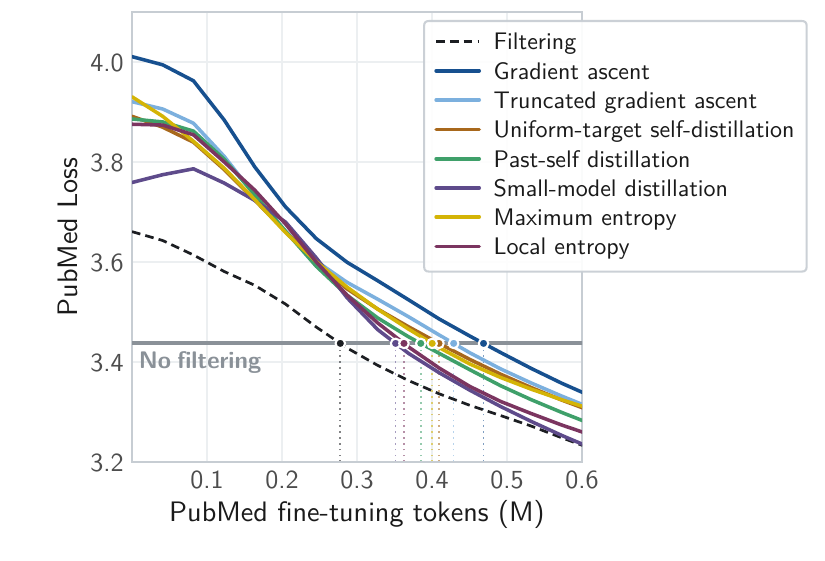}
\end{center}
\caption{
\textbf{Gradient ascent slows fine-tuning on PubMed the most among the suppression objectives.} We fine-tune each 113M \texttt{AdamW} model on PubMed and report held-out PubMed loss. The horizontal line marks the PubMed loss of \textsc{No filtering}. Results are for one fine-tuning seed.
}
\label{fig:pubmed_curves}
\end{figure}

\subsection{Suppression hyperparameters}
\label{app:hparams}
Table~\ref{tab:hparams} lists $t_{\mathrm{start}}$, as a share of pretraining, $P$ and $c$ for every suppression-pretrained model. Every model uses $S = 1$ without a retain loss and pretrains on $\mathcal{D}_\rho$, except the Unfiltered corpus models, which pretrain on $\mathcal{D}$.

\begin{table}[H]
\caption{Suppression hyperparameters.}
\label{tab:hparams}
\begin{center}
\small
\setlength{\tabcolsep}{4pt}
\begin{tabular}{@{}>{\raggedright\arraybackslash}p{3.95cm} l c c l >{\raggedright\arraybackslash}p{2.65cm}@{}}
\toprule
 & Model & Start & $P$ & $c$ & Reported in \\
\midrule
\multicolumn{6}{@{}l}{\textit{\texttt{Muon}, gradient ascent, whole document}} \\
Main models              & 113M, 224M, 521M & 0\%  & 50 & 5   & Tables~\ref{tab:resist}, \ref{tab:start_interval}, \ref{tab:positions}, \ref{tab:attack_pubmed}; Figures~\ref{fig:scale_losses}--\ref{fig:fig_frontier60}, \ref{fig:ga_evals} \\
                         & 1030M & 0\%  & 50 & 3 & Tables~\ref{tab:resist}, \ref{tab:posthoc_521m}, \ref{tab:attack_pubmed}; Figures~\ref{fig:scale_losses}--\ref{fig:scale_tokens}, \ref{fig:ga_evals} \\
Start time               & 113M & 20\% & 40 & 4.5 & Table~\ref{tab:start_interval} \\
                         & 113M & 60\% & 20 & 3.8 & Tables~\ref{tab:start_interval}, \ref{tab:p20_start} \\
                         & 113M & 0\%  & 20 & 2.2 & Table~\ref{tab:p20_start} \\
                         & 224M & 20\% & 40 & 4.2 & Table~\ref{tab:start_interval} \\
                         & 224M & 60\% & 20 & 3.8 & Table~\ref{tab:start_interval} \\
                         & 521M & 60\% & 20 & 3.0 & Table~\ref{tab:start_interval} \\
Suppression strength     & 113M & 60\% & 20 & 2.5, 3.8, 5, 6    & Figures~\ref{fig:fig_frontier60}, \ref{fig:fig_113_frontier} \\
                         & 224M & 60\% & 20 & 1.9, 2.9, 3.8, 4.8 & Figure~\ref{fig:fig_frontier60} \\
                         & 521M & 60\% & 20 & 2, 3, 3.7, 5          & Figure~\ref{fig:fig_frontier60} \\
Unfiltered corpus        & 113M & 60\% & 20 & 3.3, 5.4, 6.3, 7.8 & Figure~\ref{fig:fig_113_frontier} \\
Token-level              & 113M & 0\%  & 50 & 5   & Table~\ref{tab:positions} \\
Separate and shared statistics & 113M & 5\% & 50 & 2 & Table~\ref{tab:optimizer} \\
\midrule
\multicolumn{6}{@{}l}{\textit{\texttt{AdamW}, gradient ascent, whole document}} \\
Main models              & 113M & 5\%  & 50 & 2   & Tables~\ref{tab:optimizer}, \ref{tab:positions}, \ref{tab:attack_pubmed} \\
                         & 224M & 5\%  & 50 & 3   & Tables~\ref{tab:optimizer}, \ref{tab:attack_pubmed} \\
Shared statistics        & 113M & 5\%  & 50 & 2   & Table~\ref{tab:optimizer} \\
                         & 224M & 5\%  & 50 & 4   & Table~\ref{tab:optimizer} \\
\midrule
\multicolumn{6}{@{}l}{\textit{\texttt{AdamW}, token-level}} \\
Gradient ascent          & 113M & 5\%  & 50 & 2, 2.5, 3, 4 & Table~\ref{tab:positions}; Figures~\ref{fig:fig_pubmed_resist}, \ref{fig:pubmed_curves} \\
                         & 224M & 5\%  & 50 & 4   & Figure~\ref{fig:fig_pubmed_resist} \\
Truncated gradient ascent & 113M & 5\% & 50 & 2, 3 & Figures~\ref{fig:fig_pubmed_resist}, \ref{fig:pubmed_curves} \\
Uniform-target self-distillation & 113M & 5\% & 50 & 2, 3 & Figures~\ref{fig:fig_pubmed_resist}, \ref{fig:pubmed_curves} \\
Past-self distillation   & 113M & 5\%  & 50 & 2, 3 & Figures~\ref{fig:fig_pubmed_resist}, \ref{fig:pubmed_curves} \\
                         & 224M & 5\%  & 50 & 4   & Figure~\ref{fig:fig_pubmed_resist} \\
\quad top 1,024 tokens   & 113M & 5\%  & 50 & 2   & Tables~\ref{tab:past_self_checkpoint}, \ref{tab:retain} \\
\quad teacher at initialization & 113M & 5\% & 50 & 2 & Table~\ref{tab:past_self_checkpoint} \\
\quad rolling teacher & 113M & 5\% & 50 & 2 & Table~\ref{tab:past_self_checkpoint} \\
Small-model distillation & 113M & 5\% & 50 & 2, 3 & Figures~\ref{fig:fig_pubmed_resist}, \ref{fig:pubmed_curves} \\
Maximum entropy          & 113M & 5\%  & 50 & 2   & Figures~\ref{fig:fig_pubmed_resist}, \ref{fig:pubmed_curves} \\
Local entropy            & 113M & 5\%  & 50 & 2   & Figures~\ref{fig:fig_pubmed_resist}, \ref{fig:pubmed_curves} \\
\bottomrule
\end{tabular}
\end{center}
\end{table}

\clearpage

\section{Data}
\label{app:data}

\subsection{Target domain}
\label{app:domain}
\paragraph{Category definition.} We target medical content, following \citet{rathi2026unlearning}. A document is medical when its subject is clinical information, symptoms, diagnoses or treatments; medical or pharmaceutical research or industry; medical devices and procedures; human anatomy and physiology; virology, immunology or disease in humans; neurology and neurological disorders; or medical genetics. It is not medical when its subject is colloquial, non-medical references to anatomy; cosmetic surgery; general non-medical biology or chemistry; animal behaviour and cognition; non-medical biochemistry or genetics; healthcare policy, administration or education; psychiatry, mental illness or psychology; wellness and meditation; public health and epidemiology; or pregnancy and childcare. The document annotator chooses one of five classes --- medical, veterinary, molecular biology, organismal biology, other --- and we take medical against the remaining four, which keeps animal medicine and both biologies off the medical label. Both lists appear verbatim in every annotation prompt, and the document prompt directs the annotator to judge a document's subject rather than its vocabulary.

\paragraph{Document annotation.} The annotator is Qwen3-30B-A3B-Instruct-2507, taking a majority vote over five samples. It labels 50,000 randomly sampled FineWeb-Edu documents, which we split by document into training, development and test sets. On 300 held-out documents that we labelled by hand, the annotator reaches an $F_1$ of 0.89.

\subsection{Document classifier}
\paragraph{Document classifier.} We train a logistic-regression classifier on frozen \texttt{bge-base-en-v1.5} representations. Documents exceed the encoder's context window, so we embed each document in chunks that fill it, $L_2$-normalise each chunk's \texttt{[CLS]} embedding, and average the chunks into one document embedding, which keeps every token of a document inside its representation. We select the classification threshold by maximizing $F_1$ on the development split and apply it unchanged to the test split, where the classifier achieves an $F_1$ of 0.78 and an AUROC of 0.98.

\subsection{Forget set $\mathcal{D}_F$}
\label{app:sets}
\paragraph{Selecting the medical (target-domain) documents.} $\mathcal{D}_F$ consists of the highest-scoring documents that together account for 5\% of corpus tokens, which sets its threshold at 0.75. We set this cut for precision: suppression trains against every document in $\mathcal{D}_F$, so a general document admitted to it becomes a negative example for retained capabilities. On the test split, the threshold reaches a precision of 0.94 at a recall of 0.51.

\paragraph{Selecting the tokens.}
\label{app:mask}
Token-level suppression applies its loss only to the tokens of $\mathcal{D}_F$ selected by the token mask $m$. The default mask keeps tokens that the \textsc{Token filtering} probe (Appendix~\ref{app:baselines}) labels medical at its $F_1$-maximizing threshold. On a representative sample of $\mathcal{D}_F$-like documents, the probe's training labels mark 71\% of tokens as medical, and the $F_1$ mask selects 73\%. We therefore use this as the default token-level mask. We additionally evaluate a more conservative \emph{confident} mask that keeps only the 20\% of tokens receiving the highest medical scores. We choose 20\% as a deliberately restrictive operating point to test whether concentrating suppression on the highest-confidence tokens reduces spillover to retained content.
\clearpage

\section{Evaluation}
\label{app:eval}

\subsection{Baselines}
\label{app:baselines}
\paragraph{Filtering.} The classifier of Appendix~\ref{app:data} scores every document, and we remove the highest-scoring documents until they account for $\rho = 20\%$ of corpus tokens. We set this cut for recall: on the test split, it removes 99\% of medical documents at a precision of 0.45. At the $F_1$-maximizing threshold, 15\% of medical documents would remain in the pretraining corpus. Medical documents that pass the filter stay available throughout pretraining, and a model's target-domain loss under filtering falls quickly when even a few target-domain documents pass the filter \citep{shilov2025sgtm}. \citet{obrien2026deepignorance} likewise set their filters to minimize false negatives, and \citet{rathi2026unlearning} find that trading precision for recall makes even weak classifiers effective filters given enough pretraining compute, at a cost to performance across the board.

\paragraph{Token filtering.} We follow the token-level filtering of \citet{rathi2026unlearning}. A linear probe on the concatenated layer-13 representations of a left-to-right and a right-to-left 224M language model, both pretrained on FineWeb-Edu, scores every token; end-of-document tokens are never hidden. We train the probe on token labels derived from sparse-autoencoder latents of Gemma~2~9B as in their work, keeping medical latents consistent with the category definition of Appendix~\ref{app:domain} and requiring three active latents per medical token. We parameterize the filtering threshold by the share of corpus tokens hidden, following \citet{rathi2026unlearning}. To examine the trade-off beyond the matched-loss comparison of Table~\ref{tab:resist}, at 113M we additionally evaluate 9.2\%, 20\%, and 50\% filtering, where 9.2\% is the $F_1$-maximizing threshold (Figure~\ref{fig:fig_113_frontier}).

\paragraph{Post-hoc baselines.} Following \citet{li2024wmdp}, we take the loss at the middle layer and update that layer and the two preceding it, targeting MLP layers only. \textsc{RMU} scales the random vector by a coefficient $c$. Activation norms differ across models and optimizers, so the same $c$ is a different intervention in each; we therefore set $c = \beta \lVert h \rVert$, where $\lVert h \rVert$ is the mean activation norm at the steered layer, and tune and report $\beta$. We give \textsc{RMU} the same medical-data budget as suppression during pretraining: across all \textsc{RMU} updates, the forget batches contain the same total number of FineWeb-Edu tokens from $\mathcal{D}_F$ as were processed across all suppression steps during pretraining. Each step additionally draws an equal-sized retain batch from $\mathcal{D}_\rho$; the retain term penalizes the squared distance between the middle-layer activations of the updated model and those of the frozen starting model, weighted by $\alpha$. We select $\alpha$, $\beta$, the learning rate and the steered layers by grid search; Table~\ref{tab:rmu_grid} lists the values searched and used. For each model size, we select the \textsc{RMU} configuration requiring the most PubMed fine-tuning tokens among those with broadly comparable retained performance, allowing greater general-loss degradation than suppression. The selected configurations have comparable general task loss across scales (Table~\ref{tab:resist}). General loss is measured on held-out FineWeb-Edu documents that pass the $\rho=20\%$ filter, relative to \textsc{No filtering}. Gradient ascent \citep{jang2023knowledge}, \textsc{SimNPO} \citep{fan2025simnpo} and \textsc{Circuit Breakers} \citep{zou2024circuitbreakers}, run at matched general loss with \texttt{AdamW}, resisted less, so we report \textsc{RMU} alone. Gradient ascent is the post-hoc counterpart of suppression during pretraining: it applies the same loss to the same documents $\mathcal{D}_F$, but to the finished model, with a KL term to the model's starting predictions on $\mathcal{D}_\rho$ in place of the interleaved pretraining steps. We report it at 521M in Table~\ref{tab:posthoc_521m} for completeness.

\begin{table}[H]
\caption{\textsc{RMU} hyperparameters.}
\label{tab:rmu_grid}
\begin{center}
\fontsize{8}{9.6}\selectfont
\setlength{\tabcolsep}{3pt}
\begin{tabular}{@{}l >{\raggedright\arraybackslash}p{3.15cm} >{\raggedright\arraybackslash}p{3.25cm} l l l l@{}}
\toprule
 & \multicolumn{2}{c}{Searched} & \multicolumn{4}{c}{Used} \\
\cmidrule(lr){2-3}\cmidrule(l){4-7}
 & \texttt{AdamW} & \texttt{Muon} & 113M & 224M & 521M & 1030M \\
\midrule
Retain weight $\alpha$    & 0.1, 1, 2, 5, 10, 50, 100, 1000, 1200 & 2, 4, 12, 25, 50, 100, 125, 150, 175, 250, 500, 1000 & 125 & 50 & 25 & 1 \\
Steering strength $\beta$ & 0.5, 2.5, 5, 7.5, 10, 12.5, 25 & 5, 12.5 & 5 & 5 & 5 & 5 \\
Learning rate             & $3{\times}10^{-6}$; $3,4,5,6{\times}10^{-5}$; $1,2,3,6{\times}10^{-4}$; $1,2{\times}10^{-3}$ & $4{\times}10^{-5}$; $1,1.15,1.5{\times}10^{-4}$; $3,4,6{\times}10^{-4}$; $1,3{\times}10^{-3}$ & $1{\times}10^{-3}$ & $4{\times}10^{-4}$ & $3{\times}10^{-4}$ & $1.5{\times}10^{-4}$ \\
Steered activation        & MLP, block & MLP & MLP & MLP & MLP & MLP \\
\bottomrule
\end{tabular}
\end{center}
\end{table}

\paragraph{Post-hoc optimizer choice.} Our main post-hoc experiments use the optimizer with which each model was pretrained. Since optimizer choice can affect post-hoc unlearning \citep{zhong2025dualoptim}, we repeat the post-hoc baselines with \texttt{AdamW} for our two largest models, which are pretrained with \texttt{Muon} (gradient ascent at 521M only). Switching to \texttt{AdamW} does not increase resistance for either baseline (Table~\ref{tab:posthoc_521m}), indicating that their limited resistance is not explained by the post-hoc optimizer.

\begin{table}[H]
\caption{
\textbf{Changing the post-hoc optimizer does not increase resistance.} General loss is measured on held-out FineWeb-Edu documents that pass the 20\% filter, and PubMed loss on the held-out PubMed split; both are reported before fine-tuning and relative to \textsc{No filtering}. The final column gives the number of PubMed fine-tuning tokens required to reach the PubMed loss of \textsc{No filtering}, normalized by \textsc{Filtering}. Results are averaged over three fine-tuning seeds. \textsc{Deter} denotes suppression during pretraining. $^\dagger$1030M settings: $c = 3$.
}

\label{tab:posthoc_521m}
\centering
\small\renewcommand{\arraystretch}{0.9}
\begin{tabular}{@{}lllccc@{}}
\toprule
Size & Method & Optimizer
& \shortstack{General\\loss $\downarrow$}
& \shortstack{PubMed\\loss}
& \shortstack{Fine-tuning tokens\\$\div$ \textsc{Filtering} $\uparrow$} \\
\midrule
521M
& \textsc{Filtering}    & \texttt{Muon}   & $-0.02$ & $+0.23$  & $1.00\times$ \\
& Gradient ascent       & \texttt{Muon}   & $+0.05$ & $+0.47$  & $1.06\times$ \\
& Gradient ascent       & \texttt{AdamW}  & $+0.07$ & $+0.52$  & $1.04\times$ \\
& \textsc{RMU}          & \texttt{Muon}   & $+0.04$ & $+3.28$  & $1.40\times$ \\
& \textsc{RMU}          & \texttt{AdamW}  & $+0.04$ & $+3.18$  & $1.34\times$ \\
& \deter                & \texttt{Muon}   & $+0.04$ & $+11.10$ & $\mathbf{13.53\times}$ \\
\midrule
1030M$^\dagger$
& \textsc{Filtering}    & \texttt{Muon}   & $-0.02$ & $+0.21$  & $1.00\times$ \\
& \textsc{RMU}          & \texttt{Muon}   & $+0.06$ & $+4.05$  & $1.47\times$ \\
& \textsc{RMU}          & \texttt{AdamW}  & $+0.04$ & $+4.23$  & $1.45\times$ \\
& \deter                & \texttt{Muon}   & $+0.01$ & $+8.04$  & $\mathbf{20.36\times}$ \\
\bottomrule
\end{tabular}
\end{table}

\subsection{Corpus evaluation sets}
\label{app:corpussets}

For the bioRxiv evaluation\footnote{\url{https://huggingface.co/datasets/mteb/raw_biorxiv}}, we group eight author-assigned subjects as medical-adjacent: neuroscience, immunology, microbiology, cancer biology, physiology, pharmacology and toxicology, pathology, and epidemiology. The remaining 18 subjects form the biology set.

The PubMed evaluation set is drawn from a Common Pile \citep{kandpal2025commonpile} shard disjoint from those used for adversarial fine-tuning.

\subsection{Task loss}
\label{app:taskloss}

\begin{wraptable}{r}{0.59\linewidth}
\caption{Benchmarks used for task loss.}
\label{tab:benchmarks}
\centering
\small
\setlength{\tabcolsep}{3pt}
\begin{tabular}{@{}llr@{}}
\toprule
Benchmark & Split / subset & Items \\
\midrule
\multicolumn{3}{@{}l}{\emph{Medical}} \\
\hspace{0.5em}MedQuAD & Answered questions & 16,407 \\
\hspace{0.5em}MedMCQA & Validation & 4,183 \\
\hspace{0.5em}MedQA-USMLE (4 options) & Test & 1,273 \\
\hspace{0.5em}MMLU medicine (6 subjects) & Test & 1,111 \\
\addlinespace[2pt]
\multicolumn{3}{@{}l}{\emph{Biology}} \\
\hspace{0.5em}MMLU biology (2 subjects) & Test & 454 \\
\addlinespace[2pt]
\multicolumn{3}{@{}l}{\emph{General}} \\
\hspace{0.5em}MMLU STEM (16 subjects) & Test & 2,564 \\
\hspace{0.5em}MMLU non-STEM (30 subjects) & Test & 9,253 \\
\bottomrule
\end{tabular}
\end{wraptable}

We score every benchmark using the cloze formulation of \citet{rathi2026unlearning} for their base models: the model reads \texttt{Question:}, followed by the question, a newline, and \texttt{Answer:}, and we compute the negative log-likelihood of the correct answer text, prefixed by a space. We include no answer choices or demonstrations. We average the negative log-likelihood over the tokens of the answer. All our models share one tokenizer, so we normalize by tokens; \citet{bhagia2025establishing} instead normalize by bytes to reduce the effect of differing tokenizers. We first average over items within each benchmark, then weight benchmarks equally within each group: medical (MedQuAD, MedMCQA, MedQA-USMLE and MMLU medicine), biology (MMLU biology) and general (MMLU STEM and MMLU non-STEM). MMLU medicine consists of the six subjects used by \citet{rathi2026unlearning}: anatomy, clinical knowledge, college medicine, medical genetics, professional medicine and virology. MMLU biology consists of college and high school biology, MMLU STEM of the remaining subjects in MMLU's STEM category, and MMLU non-STEM of all other subjects except human aging, nutrition and human sexuality, which concern health and are excluded from all three groups. We use the MedMCQA validation set because its test labels are not public. For the 114 MedQuAD items whose question--answer sequence exceeds the 2,048-token context, we truncate the answer to fit; every other item fits. Table~\ref{tab:benchmarks} lists the evaluation splits and number of items.

To reduce sensitivity to checkpoint-to-checkpoint variation, we average task loss over four checkpoints from the final 10\% of training, following \citet{magnusson2025datadecide}. At each checkpoint, we subtract the task loss of \textsc{No filtering} at the corresponding training step and report the mean difference and standard deviation across checkpoints. For \textsc{RMU}, we apply the same edit with the same hyperparameters independently to the corresponding \textsc{Filtering} checkpoint. Corpus loss and adversarial fine-tuning use the final checkpoint.

\subsection{Adversarial fine-tuning}
\label{app:attack}

\begin{wraptable}{r}{0.525\linewidth}
\caption{Selected adversary hyperparameters on PubMed and fine-tuning tokens to parity. For \texttt{Muon} rows, tokens to parity are averaged over three fine-tuning seeds, and the listed rate is the fastest on each seed unless marked; \texttt{AdamW} rows are for one seed.}
\label{tab:attack_pubmed}
\centering
\small
\setlength{\tabcolsep}{4pt}
\begin{threeparttable}
\begin{tabular}{@{}llcc@{}}
\toprule
Model & Method & \begin{tabular}[b]{@{}c@{}}Peak\\ learning rate\end{tabular} & \begin{tabular}[b]{@{}c@{}}Tokens to\\ parity (M)\end{tabular} \\
\midrule
\multicolumn{4}{@{}l}{\textit{Pretrained with \texttt{Muon}}} \\
113M & \textsc{Filtering}   & $2{\times}10^{-3}$ & 0.13 \\
     & \deter ($c = 5$)     & $2{\times}10^{-3}$ & 0.39 \\
     & \textsc{RMU}         & $2{\times}10^{-3}$ & 0.17 \\
224M & \textsc{Filtering}   & $2{\times}10^{-3}$ & 0.12 \\
     & \deter ($c = 5$)     & $2{\times}10^{-3}$ & 0.47 \\
     & \textsc{RMU}         & $2{\times}10^{-3}$ & 0.17 \\
521M & \textsc{Filtering}   & $2{\times}10^{-3}$ & 0.12 \\
     & \deter ($c = 5$)     & $2{\times}10^{-3}$ & 1.58 \\
     & \textsc{RMU}         & $2{\times}10^{-3}$ & 0.16 \\
1030M & \textsc{Filtering}   & $10^{-3}$          & 0.10 \\
     & \deter ($c = 3$)     & $10^{-3}$\rlap{\tnote{$\ddagger$}} & 2.09 \\
     & \textsc{RMU}         & $2{\times}10^{-3}$ & 0.15 \\
\midrule
\multicolumn{4}{@{}l}{\textit{Pretrained with \texttt{Muon}, later start}} \\
113M & \deter (start 20\%) & $2{\times}10^{-3}$ & 0.38 \\
     & \deter (start 60\%) & $2{\times}10^{-3}$ & 0.47 \\
224M & \deter (start 20\%) & $2{\times}10^{-3}$\rlap{\tnote{$\dagger$}} & 0.71 \\
     & \deter (start 60\%) & $2{\times}10^{-3}$\rlap{\tnote{$\dagger$}} & 0.79 \\
521M & \deter (start 60\%) & $2{\times}10^{-3}$ & 1.83 \\
\midrule
\multicolumn{4}{@{}l}{\textit{Pretrained with \texttt{AdamW}}} \\
113M & \textsc{Filtering}   & $5{\times}10^{-4}$ & 0.27 \\
     & \deter ($c = 2$)     & $10^{-3}$          & 0.53 \\
     & \textsc{RMU}         & $5{\times}10^{-4}$ & 0.32 \\
224M & \textsc{Filtering}   & $5{\times}10^{-4}$ & 0.19 \\
     & \deter ($c = 3$)     & $10^{-3}$          & 0.64 \\
     & \textsc{RMU}         & $10^{-3}$          & 0.29 \\
\bottomrule
\end{tabular}
\begin{tablenotes}\footnotesize
\item[$\dagger$] $5{\times}10^{-3}$ on one of the three seeds.
\item[$\ddagger$] $2{\times}10^{-3}$ on one of the three seeds.
\end{tablenotes}
\end{threeparttable}
\end{wraptable}

\paragraph{The adversary.} The adversary fine-tunes the final pretraining checkpoint with the optimizer the model was pretrained with. A mismatch between the fine-tuning and pretraining optimizers weakens fine-tuning \citep{liu2025moonlight, qu2026muon}, and against our \texttt{Muon} models an \texttt{AdamW} adversary is slower. Following \citet{rathi2026unlearning}, gradients are clipped at norm $0.5$, each step takes 40,960 tokens as 20 sequences of 2,048, and the moment estimates start fresh. With \texttt{Muon}, momentum is $0.95$ and all parameter groups share one peak rate. The fine-tuning seed sets the order of the PubMed fine-tuning data; each model is pretrained once.

\paragraph{Adversarial Fine-tuning on PubMed.} The adversary fine-tunes on the PubMed section of the Common Pile \citep{kandpal2025commonpile} for up to 100 steps, taking one pass without replacement, and we score medical log-perplexity after every step on a held-out PubMed validation split. For our two largest models, we continue the attack to 0.08\% of pretraining (8M tokens at 521M and 16M at 1030M) to test whether the gap to \textsc{Filtering} persists beyond this point. The target is parity, \textsc{No filtering}'s own loss on that split, which every method reaches inside the budget (Figures~\ref{fig:pubmed_curves_atkmuon} and~\ref{fig:pubmed_curves}). We use weight decay $0.01$. For an adversary with \texttt{AdamW}, we search over rates from $\{10^{-5},\allowbreak 5{\times}10^{-5},\allowbreak 10^{-4},\allowbreak 3{\times}10^{-4},\allowbreak 5{\times}10^{-4},\allowbreak 10^{-3},\allowbreak 2{\times}10^{-3},\allowbreak 3{\times}10^{-3},\allowbreak 5{\times}10^{-3}\}$ and warmups from $\{0,\allowbreak 3,\allowbreak 6,\allowbreak 10,\allowbreak 40\}$ steps. With \texttt{Muon}, we search over rates from $\{3{\times}10^{-4},\allowbreak 10^{-3},\allowbreak 2{\times}10^{-3},\allowbreak 5{\times}10^{-3},\allowbreak 10^{-2}\}$ and warmups from $\{0,\allowbreak 3,\allowbreak 10\}$ steps. For each model and fine-tuning seed we select the learning rate that reaches parity in the fewest steps (Table~\ref{tab:attack_pubmed}). In our warmup ablations, a 3-step warmup reaches parity fastest with \texttt{AdamW}, and no warmup reaches parity fastest with \texttt{Muon}. For every model, the selected rate lies inside the searched range: the next lower and next higher rates both reach parity more slowly or not at all. We report resistance as the ratio of a model's fine-tuning tokens to parity to those of the filtered baseline.

\clearpage
\section{Additional results with \texttt{Muon}}
\label{app:results}

\begin{figure}[H]
\begin{center}
\includegraphics[width=\linewidth]{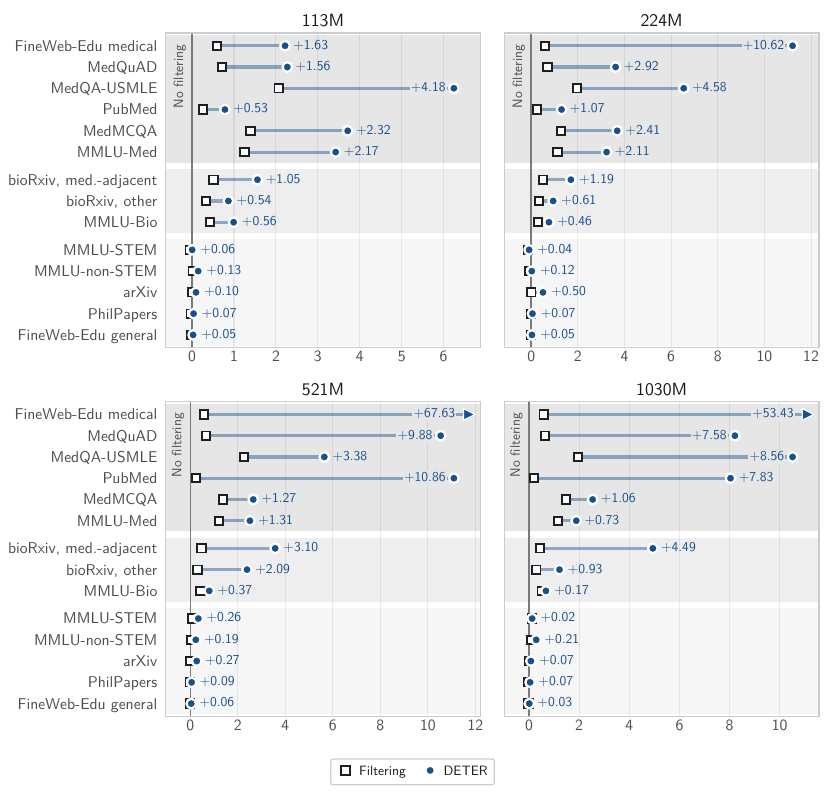}
\end{center}
\caption{
\textbf{Suppression primarily affects medical evaluations, with smaller effects on biology and general-domain evaluations.} We report log-perplexity relative to \textsc{No filtering}. Squares show \textsc{Filtering} and circles show suppression during pretraining; annotations give the increase from \textsc{Filtering} to suppression for each evaluation.
}
\label{fig:ga_evals}
\end{figure}

\paragraph{Effect on each evaluation.} Figure~\ref{fig:ga_evals} breaks down the increase in loss from suppression relative to \textsc{Filtering} for each evaluation set. Across model sizes, the increase is consistently larger on medical evaluations than on general-domain evaluations. Among the biology evaluations, the medically adjacent bioRxiv split is affected most at every size.

\begin{figure}[t]
\centering
\includegraphics[width=0.8\linewidth,trim=0 7bp 0 0,clip]{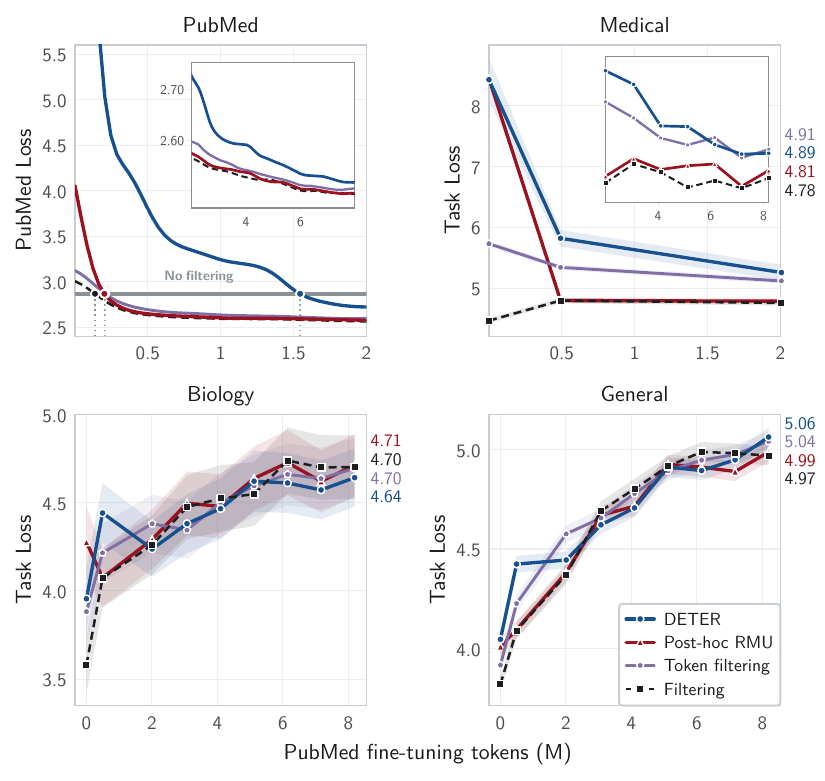}
\caption{
\textbf{At 521M, post-hoc \textsc{RMU} returns to the medical task loss of \textsc{Filtering} early in PubMed fine-tuning, while suppression during pretraining keeps medical task loss higher for longer.} We fine-tune each model on PubMed and report held-out PubMed loss and task loss on the medical benchmarks (MedQuAD, MedMCQA, MedQA-USMLE, and MMLU medicine), MMLU biology, and MMLU STEM and non-STEM (general). The horizontal line marks the PubMed loss of \textsc{No filtering}. Numbers at the right give the task loss at the end of fine-tuning. Results are for one fine-tuning seed; shaded bands are 95\% bootstrap intervals over benchmark items. \textsc{Deter} denotes suppression during pretraining.}
\label{fig:bench_recovery_521m}
\end{figure}

\paragraph{What remains on downstream tasks after reaching the \textsc{No filtering} PubMed loss?} Because our resistance metric is defined by held-out PubMed loss, we also measure downstream task loss when that criterion is first met. For each method, we evaluate the first checkpoint whose PubMed loss matches \textsc{No filtering} (Table~\ref{tab:task_loss_parity}). These checkpoints occur after different amounts of fine-tuning, so the table is not a matched-compute comparison; Figure~\ref{fig:bench_recovery_521m} shows the task-loss trajectories at matched fine-tuning budgets for the 521M model.

\begin{table}[H]
\caption{
\textbf{Task losses before fine-tuning and when each method first reaches the PubMed loss of \textsc{No filtering}.} We fine-tune each model on PubMed until its held-out PubMed loss first reaches that of \textsc{No filtering}, and report task loss at the final pretraining checkpoint (before) and at that checkpoint (after). All methods in the right block are therefore evaluated at the same PubMed loss, but reach it after different amounts of fine-tuning, reported as PubMed fine-tuning tokens. \textbf{The right block is not a matched-compute comparison:} each method is evaluated at its own checkpoint where PubMed loss first reaches that of \textsc{No filtering}. Figure~\ref{fig:bench_recovery_521m} instead compares task loss after equal amounts of PubMed fine-tuning at 521M. Medical task loss is averaged over MedQuAD, MedMCQA, MedQA-USMLE, and MMLU medicine; general task loss is measured on MMLU STEM and non-STEM; and closest-retain task loss on MMLU biology. All task losses are reported relative to the final checkpoint of \textsc{No filtering}. Results are averaged over three fine-tuning seeds. \textsc{Deter} denotes suppression during pretraining.
}
\small
\label{tab:task_loss_parity}
\centering
\setlength{\tabcolsep}{2.5pt}
\setlength{\belowrulesep}{0.4ex}
\begin{tabular}{@{}llccccccc@{}}
\toprule
& &
\multicolumn{3}{c}{\shortstack{Before medical\\fine-tuning on PubMed}} &
\multicolumn{4}{c}{\shortstack{At first checkpoint reaching the\\
PubMed loss of \textsc{No filtering}}} \\
\cmidrule(lr){3-5}
\cmidrule(l){6-9}
Size & Method
& \shortstack{General\\task loss $\downarrow$}
& \shortstack{Closest-\\retain task\\loss $\downarrow$}
& \shortstack{Medical\\task loss $\uparrow$}
& \shortstack{PubMed FT\\tokens (M)}
& \shortstack{General\\task loss}
& \shortstack{Closest-\\retain task\\loss}
& \shortstack{Medical\\task loss} \\
\midrule

113M
& \textsc{Filtering}       & $-0.02$ & $+0.43$ & $+1.36$ & $0.13$ & $+0.11$ & $+0.59$ & $+1.39$ \\
& \textsc{Token filtering} & $+0.13$ & $+0.77$ & $+2.40$ & $0.16$ & $+0.27$ & $+0.84$ & $+2.14$ \\
& \textsc{RMU}             & $+0.13$ & $+0.92$ & $+3.58$ & $0.17$ & $+0.15$ & $+0.69$ & $+1.54$ \\
& \deter                   & $+0.07$ & $+0.99$ & $+3.91$ & $0.39$ & $+0.40$ & $+1.13$ & $+2.84$ \\

\midrule

224M
& \textsc{Filtering}       & $-0.11$ & $+0.30$ & $+1.26$ & $0.12$ & $0.00$ & $+0.47$ & $+1.32$ \\
& \textsc{Token filtering} & $+0.02$ & $+0.69$ & $+2.54$ & $0.18$ & $+0.18$ & $+0.83$ & $+2.29$ \\
& \textsc{RMU}             & $-0.03$ & $+0.78$ & $+3.79$ & $0.17$ & $+0.01$ & $+0.50$ & $+1.43$ \\
& \deter                   & $-0.03$ & $+0.76$ & $+4.27$ & $0.47$ & $+0.26$ & $+0.87$ & $+2.71$ \\

\midrule

521M
& \textsc{Filtering}       & $+0.06$ & $+0.42$ & $+1.37$ & $0.12$ & $+0.15$ & $+0.66$ & $+1.54$ \\
& \textsc{Token filtering} & $+0.15$ & $+0.73$ & $+2.64$ & $0.18$ & $+0.31$ & $+0.91$ & $+2.34$ \\
& \textsc{RMU}             & $+0.24$ & $+1.12$ & $+5.35$ & $0.16$ & $+0.16$ & $+0.67$ & $+1.63$ \\
& \deter                   & $+0.28$ & $+0.80$ & $+5.33$ & $1.58$ & $+0.71$ & $+1.18$ & $+2.32$ \\

\bottomrule
\end{tabular}
\end{table}

\paragraph{Suppression interval and start time.} Table~\ref{tab:start_interval} keeps the number of suppression steps fixed across onsets, so later starts suppress more often. To separate the interval from the onset, we also suppress from the start of pretraining at 113M with the same interval as the 60\% start. Although this model takes more suppression steps at approximately the same general loss, it requires less target-domain fine-tuning to reach the \textsc{No filtering} loss (Table~\ref{tab:p20_start}). The shorter interval therefore does not account for the advantage of starting suppression later.

\begin{table}[H]
\caption{\textbf{Using the shorter interval $P = 20$ from the start of pretraining requires less target-domain fine-tuning to undo than starting at 60\%.} General loss is measured on held-out retained FineWeb-Edu, relative to \textsc{No filtering}. The final column gives the number of PubMed fine-tuning tokens needed to reach the PubMed loss of \textsc{No filtering}, normalized by the corresponding value for \textsc{Filtering} at the same model size. Results are averaged over three fine-tuning seeds.}
\label{tab:p20_start}
\centering
\begin{tabular}{@{}lcc@{}}
\toprule
& Before fine-tuning & \shortstack{Target-domain\\fine-tuning} \\
\cmidrule(lr){2-2}\cmidrule(l){3-3}
Start
& \shortstack{General\\loss $\downarrow$}
& \shortstack{Fine-tuning\\tokens $\div$\\\textsc{Filtering} $\uparrow$} \\
\midrule
\multicolumn{3}{@{}l}{\textit{113M}, $P = 20$} \\
0\%  & $+0.03$ & $2.77\times$ \\
60\% & $+0.03$ & $\mathbf{3.67\times}$ \\
\bottomrule
\end{tabular}
\end{table}

\end{document}